\documentclass[10pt,twocolumn,letterpaper]{article}

\usepackage[accsupp]{axessibility}
\usepackage{cvpr}              

\usepackage{graphicx}
\usepackage{amsmath}
\usepackage{amssymb}
\usepackage{booktabs}
\usepackage{lineno}
\usepackage{indentfirst} 
\usepackage{commath}
\usepackage{tabularx}
\usepackage{float}
\usepackage{stfloats}
\usepackage{algorithm}
\usepackage{arydshln}
\usepackage{algpseudocode}
\usepackage{graphicx}
\usepackage{color}
\usepackage{CJK}
\usepackage{multirow}
\usepackage{multicol}
\usepackage[pagebackref,breaklinks,colorlinks]{hyperref}

\usepackage[capitalize]{cleveref}
\crefname{section}{Sec.}{Secs.}
\Crefname{section}{Section}{Sections}
\Crefname{table}{Table}{Tables}
\crefname{table}{Tab.}{Tabs.}

\def\cvprPaperID{2623} 
\def\confName{CVPR}
\def\confYear{2023}

\begin{document}

\title{3D Registration with Maximal Cliques}

\author{Xiyu Zhang \quad Jiaqi Yang\footnotemark[1] \quad Shikun Zhang \quad Yanning Zhang\\
School of Computer Science, Northwestern Polytechnical University, China\\
{\tt\small \{2426988253, zhangshikun\}@mail.nwpu.edu.cn; \{jqyang, ynzhang\}@nwpu.edu.cn}}
\maketitle
\renewcommand{\thefootnote}{\fnsymbol{footnote}}
\footnotetext[1]{Corresponding author. \\ Code will be available at \url{https://github.com/zhangxy0517/3D-Registration-with-Maximal-Cliques}.}
\renewcommand{\thefootnote}{\arabic{footnote}}
\begin{abstract}

As a fundamental problem in computer vision, 3D point cloud registration (PCR) aims to seek the optimal pose to align a point cloud pair. In this paper, we present a 3D registration method with maximal cliques (MAC). The key insight is to loosen the previous maximum clique constraint, and mine more local consensus information in a graph for accurate pose hypotheses generation: 1) A compatibility graph is constructed to render the affinity relationship between initial correspondences. 2) We search for maximal cliques in the graph, each of which represents a consensus set. We perform node-guided clique selection then, where each node corresponds to the maximal clique with the greatest graph weight. 3) Transformation hypotheses are computed for the selected cliques by the SVD algorithm and the best hypothesis is used to perform registration. Extensive experiments on U3M, 3DMatch, 3DLoMatch and KITTI demonstrate that MAC effectively increases registration accuracy, outperforms various state-of-the-art methods and boosts the performance of deep-learned methods. MAC combined with deep-learned methods achieves state-of-the-art registration recall of \textbf{95.7\% / 78.9\%} on 3DMatch / 3DLoMatch. 

\end{abstract}
\begin{figure}[t]
  \centering
  \includegraphics[width=\linewidth]{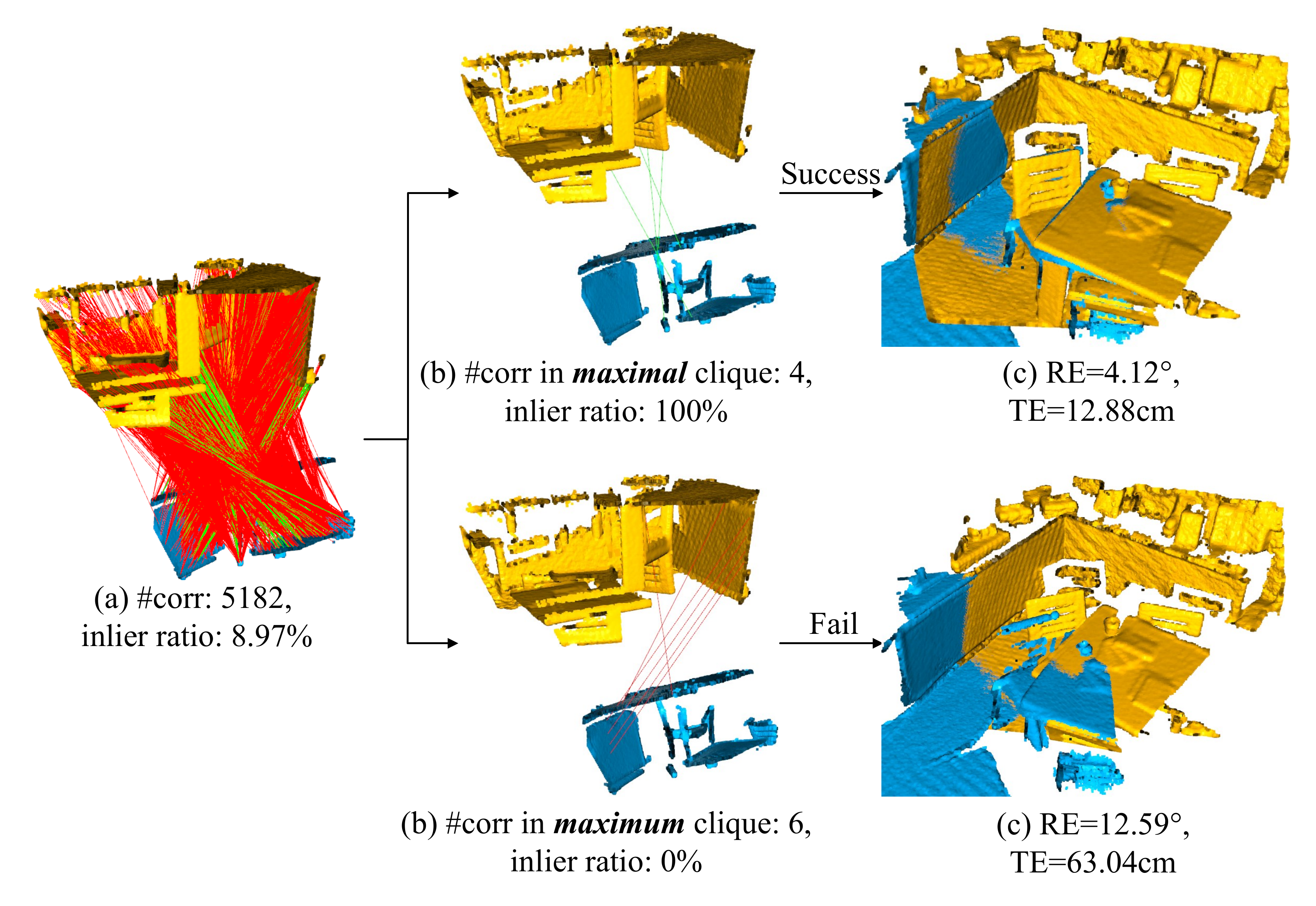}
   \caption{Comparison of {\textbf{maximal}} and {\textbf{maximum }} cliques on a low overlapping point cloud pair. Maximal cliques (MAC) effectively choose the optimal 6-DoF transformation hypothesis with low rotation error (RE) and translation error (TE) for two point clouds with a low inlier ratio, while the maximum clique fails in this case.}
   \label{fig:onecol}
\end{figure}
\section{Introduction}
\label{sec:intro}
Point cloud registration (PCR) is an important and fundamental problem in 3D computer vision and has a wide range of applications in localization~\cite{fischler1981random}, 3D object detection~\cite{guo20143d} and 3D reconstruction~\cite{mian2005automatic}. Given two 3D scans of the same object (or scene), the goal of PCR is to estimate a six-degree-of-freedom (6-DoF) pose transformation that accurately aligns the two input point clouds. Using point-to-point feature correspondences is a popular and robust solution to the PCR problem.
 However, due to the limitations of existing 3D keypoint detectors \& descriptors, the limited overlap between point clouds and data noise, correspondences generated by feature matching usually contain outliers, resulting in great challenges to accurate 3D registration. 
 
 The problem of 3D registration by handling correspondences with outliers has been studied for decades. We classify them into geometric-only and deep-learned methods. For geometric-only methods~\cite{barath2018graph,quan2020compatibility,yang2021sac,yang2021toward,yang2022correspondence,li2021practical,yang2015go,bustos2017guaranteed,rusu2009fast}, random sample consensus (RANSAC) and its variants perform an iterative sampling strategy for registration. Although RANSAC-based methods are simple and efficient, their performance  is highly vulnerable when the outlier rate increases, and it requires a large number of iterations to obtain acceptable results. Also, a series of global registration methods based on branch-and-bound (BnB) are proposed to search the 6D parameter space and obtain the optimal global solution. The main weakness of these methods is the high computational complexity, especially when the correspondence set is of a large magnitude and has an extremely high outlier rate. For deep-learned methods, some~\cite{aoki2019pointnetlk,ao2021spinnet,bai2020d3feat,choy2019fully, wang2022you, gojcic2019perfect, huang2021predator,bai2021pointdsc,pais20203dregnet,lee2021deep,choy2020deep, fu2021robust} focus on improving one module in the registration process, such as investigating more discriminate keypoint feature descriptors or more effective correspondence selection techniques, while the others~\cite{  li2022lepard,yu2021cofinet,qin2022geometric} focus on registration in an end-to-end manner. However, deep-learned based methods require a large amount of data for training and usually lack generalization on different datasets. At present, it is still very challenging to achieve accurate registrations in the presence of heavy outliers and in cross-dataset conditions.

In this paper, we propose a geometric-only 3D registration method based on  maximal cliques (MAC). The key insight is to loosen the previous maximum clique constraint, and mine more local consensus information in a graph to generate accurate pose hypotheses. We first model the initial correspondence set as a compatibility graph, where each node represents a single correspondence and each edge between two nodes indicates a pair of compatible correspondences. Second, we search for maximal cliques in the graph and then use node-guided clique filtering to match each graph node with the appropriate maximal clique containing it. {\textit{Compared with the maximum clique, MAC is a looser constraint and is able to mine more local information in a graph.}} This helps us to achieve plenty of correct hypotheses from a graph. Finally, transformation hypotheses are computed for the selected cliques by the SVD algorithm. The best hypothesis is selected to perform registration using popular hypothesis evaluation metrics in the RANSAC family. To summarize, our main contributions are as follows:
\begin{itemize}
    \item We introduce a hypothesis generation method named MAC. Our MAC method is able to mine more local information in a graph, compared with the previous maximum clique constraint. We demonstrate that hypotheses generated by MAC are of high accuracy even in the presence of heavy outliers. 
    \item Based on MAC, we present a novel PCR method, which achieves state-of-the-art performance on U3M, 3DMatch, 3DLoMatch and KITTI datasets. Notably, our geometric-only MAC method outperforms several state-of-the-art deep learning methods~\cite{choy2020deep,pais20203dregnet, bai2021pointdsc,lee2021deep}. MAC can also be inserted as a module into multiple deep-learned frameworks~\cite{choy2019fully, ao2021spinnet, huang2021predator, yu2021cofinet, qin2022geometric} to boost their performance. MAC combined with GeoTransformer achieves the state-of-the-art registration recall of \textbf{95.7\% / 78.9\%} on 3DMatch / 3DLoMatch.
\end{itemize}

\begin{figure*}[t]
\centering
    \includegraphics[width=\linewidth]{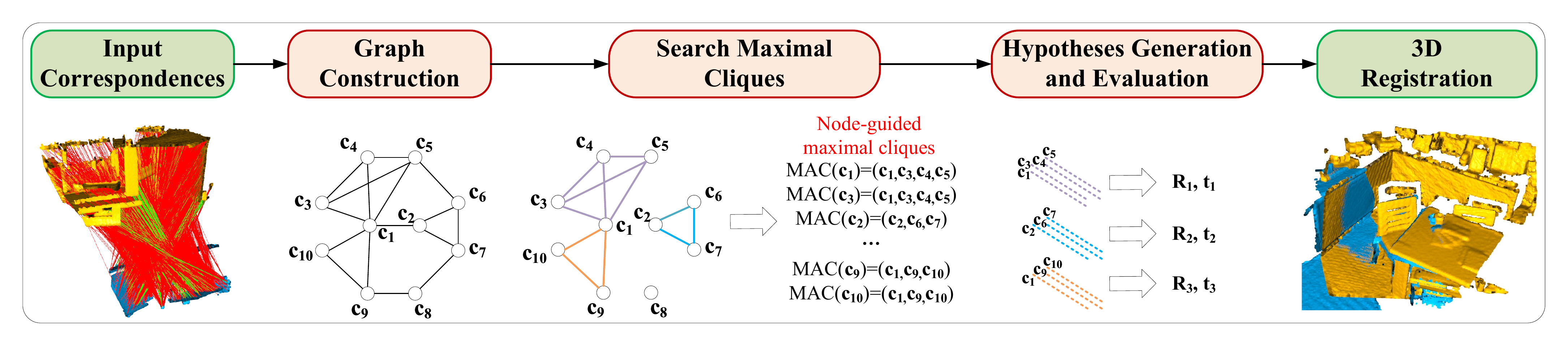}
    \caption{{\textbf{Pipeline of MAC}}. \textbf{1.} Construct a graph for the initial correspondence set. \textbf{2.} Select a set of maximal cliques from the graph as the consistent sets. \textbf{3.} Generate and evaluate the hypotheses according to the consistent sets. \textbf{4.} Select the best hypothesis to perform 3D registration.}
    \label{fig:pipline}
\end{figure*}
\section{Related Work}
\label{sec:related}
\subsection{Geometric-only PCR Methods} 
Various geometric-only methods~\cite{leordeanu2005spectral, zhou2016fast, bustos2017guaranteed, yang2020teaser, chen2022sc2} have been proposed recently. Typically, RANSAC and its variants~\cite{barath2018graph,quan2020compatibility,yang2021sac,yang2021toward,yang2022correspondence,fischler1981random,rusu2009fast} remain the dominant approaches to the problem of estimating a 6-DoF pose from correspondences. RANSAC iteratively samples correspondences from the initial set, generating and evaluating geometric estimations for each subset until a satisfactory solution is obtained. Efficient and robust evaluation metrics are extremely important for using RANSAC to achieve accurate registration. To address the current problems of time-consuming and noise-sensitive evaluation metrics, ~\cite{yang2021toward} analyzes the contribution of inliers and outliers during the computation and proposed several metrics that can effectively improve the registration performance of RANSAC. A large number of variants have also been proposed to achieve further improvement. For example, Rusu \etal~\cite{rusu2009fast} presented the simple consensus-based initial alignment (SAC-IA) method, which samples correspondences spread out on the point cloud and leverages the Huber penalty for evaluation. Graph cut RANSAC (GC-RANSAC)~\cite{barath2018graph} uses the graph-cut algorithm before model re-fitting in the local optimization step. Compatibility-guided sample consensus (CG-SAC)~\cite{quan2020compatibility} additionally considers the normal information of key points during the sampling process. Yang \etal~\cite{yang2021sac} proposed the sample consensus by sampling compatibility triangles (SAC-COT) method, which generates estimations by ranking and sampling ternary loops from the compatibility graph. Although many previous efforts have been made, these methods suffer from low time efficiency and limited accuracy in cases with high outlier rates.

A series of globally optimal methods based on BnB have been proposed recently. Yang \etal~\cite{yang2015go} proposed globally optimal ICP (GO-ICP), which rationalizes the planning of ICP update tasks at different stages, and its biggest advantage is that it minimizes the local optimum. Bustos and Chin~\cite{bustos2017guaranteed} presented guaranteed outlier removal (GORE), which calculates the tight lower bound and tight upper bound for each correspondence and reduces the size of correspondence set by rejecting true outliers. Motivated by GORE, Li\cite{li2021practical} proposed a polynomial time outlier removal method, which seeks the tight lower and upper bound by calculating the costs of correspondence matrix (CM) and augmented correspondence matrix (ACM). However, BnB techniques are sensitive to the cardinality of the input and are time-consuming for large-scale inputs.
\subsection{Deep-learned PCR Methods} 
In addition to geometric-only methods, recent works also adopt deep learning techniques to perform PCR. Some methods aim to detect more repeatable keypoints~\cite{bai2020d3feat,huang2021predator} and extract more descriptive features~\cite{choy2019fully,ao2021spinnet}. FCGF~\cite{choy2019fully} computes the features in a single pass through a fully convolutional neural network without keypoint detection. D3Feat~\cite{bai2020d3feat} uses a fully convolutional network to obtain local information of point clouds and a joint learning framework to achieve 3D local feature detection and description. Predator~\cite{huang2021predator} applies an attention mechanism to extract salient points in overlapping regions of the point clouds, thus achieving robust registration in the presence of low overlap rates. Spinnet~\cite{ao2021spinnet} extracts local features which are rotationally invariant and sufficiently informative to enable accurate registration. Some methods~\cite{choy2020deep, pais20203dregnet, bai2021pointdsc, fu2021robust} focus on efficiently distinguishing correspondences as inliers and outliers. Deep global registration (DGR)\cite{choy2020deep} and 3DRegNet\cite{pais20203dregnet} classify a given correspondence by training end-to-end neural networks and using operators such as sparse convolution and point-by-point MLP. PointDSC~\cite{bai2021pointdsc} explicitly explores spatial consistency for removing outlier correspondences and 3D point cloud registration. Fu \etal~\cite{fu2021robust} proposed a registration framework that utilizes deep graph matching (RGM) that can find robust and accurate point-to-point correspondences. More recently, several methods~\cite{yu2021cofinet,qin2022geometric} follow the detection-free methods and estimate the transformation in an end-to-end way. CoFiNet~\cite{yu2021cofinet} extracts correspondences from coarse to fine without keypoint detection. GeoTransformer~\cite{qin2022geometric} learns geometric features for robust superpoint matching and is robust in low-overlap cases and invariant to rigid transformation.

While deep learning techniques have demonstrated a great potential for PCR, these methods require a large amount of training data and their generalization is not always promising. By contrast, MAC does not require any training data and achieves more advanced performance than several deep-learned methods. Moreover, MAC can be served as a drop-on module in deep learning frameworks to boost their performance.


\section{MAC}
\label{sec:method}
\subsection{Problem Formulation}
For two point clouds ${\textbf{P}^s}$ and ${\textbf{P}^t}$ to be aligned, we first extract local features for them using geometric or learned descriptors. Let ${{\bf p}^s}$ and ${{\bf p}^t}$ denote the points in the ${\textbf{P}^s}$ and ${\textbf{P}^t}$, respectively. An initial correspondence set $\mathbf{C}_{initial}=\{{\bf c}\}$ is formed by matching feature descriptors, where ${{\bf c} = ({{\bf p}^s},{{\bf p}^t})}$. MAC estimates the 6-DoF pose transformation between ${\textbf{P}^s}$ and ${\textbf{P}^t}$ from  $\mathbf{C}_{initial}$. 

Our method is technically very simple, and its pipeline is shown in Fig.~\ref{fig:pipline}.
\subsection{Graph Construction} \label{subsec:graph}
The graph space can more accurately depict the affinity relationship between correspondences than the Euclidean space. Therefore, we model the initial correspondences as a compatibility graph, where correspondences are represented by nodes and edges link nodes that are geometrically compatible. Here, we consider two approaches to construct a compatibility graph.
\begin{itemize}
    \item {\textbf{First Order Graph.}} The first order graph (FOG) is constructed based on the rigid distance constraint between the correspondence pair $({{\bf c}_i},{{\bf c}_j})$, which can be quantitatively measured as:
    \begin{equation}\label{eq1}
        {{{S}}_{{dist}}}({{\bf c}_i},{{\bf c}_j}) = \left| {\left\| {{{\bf p}}_i^s - {{\bf p}}_j^s} \right\| - \left\| {{{\bf p}}_i^t - {{\bf p}}_j^t} \right\|} \right|.
    \end{equation}
    The compatibility score between ${{\bf c}_i}$ and ${{\bf c}_j}$ is given as:
    \begin{equation}\label{eq2}
        {{{S}}_{cmp}}({{\bf c}_i},{{\bf c}_j}) = \exp ( - \frac{{{{{S}}_{{dist}}}{{({{\bf c}_i},{{\bf c}_j})}^2}}}{{2d_{cmp}^2}}),
    \end{equation}
    where ${d_{cmp}}$ is a distance parameter. Notably, if ${{{S}}_{cmp}}({{\bf c}_i},{{\bf c}_j})$ is greater than a threshold $t_{cmp}$, ${{\bf c}_i}$ and ${{\bf c}_j}$ form an edge ${\bf e}_{ij}$ and ${{{S}}_{cmp}}({{\bf c}_i},{{\bf c}_j})$ is the weight of ${\bf e}_{ij}$, otherwise ${{{S}}_{cmp}}({{\bf c}_i},{{\bf c}_j})$ will be set to 0. Since the compatibility graph is undirected, the weight matrix ${{\mathbf{W}}_{{FOG}}}$ is symmetric.

\item{\textbf{Second Order Graph.}} The previous study~\cite{chen2022sc2} proposes a second order compatibility measure, which relates to the number of commonly compatible correspondences in the global set. The second order graph (SOG) evolves from FOG. The weight matrix ${\mathbf{W}}_{SOG}$ can be calculated as:
\begin{equation}\label{eq4}
    {{\mathbf{W}}_{SOG}} = {{\mathbf{W}}_{{FOG}}} \odot ({{\mathbf{W}}_{{FOG}}} \times {{\mathbf{W}}_{FOG}}),
\end{equation}
where $\odot$ represents the element-wise product between two matrices.
\end{itemize}

Both graph construction methods can adapt to our frameworks. Compared with FOG, 1) SOG has stricter edge construction conditions and a higher degree of compatibility with adjacent nodes; 2) SOG is sparser, which facilitates a more rapid search of cliques.
In Sec.~\ref{sec:ablation}, we experimentally compare FOG and SOG in our MAC framework. 

\subsection{Search Maximal Cliques}
Given an undirected graph $G=({\bf V},{\bf E})$, clique $C=({\bf V'},{\bf E'})$, ${\bf V'}\subseteq{\bf V}$, ${\bf E'}\subseteq{\bf E}$ is a subset of $G$, in which any two nodes are connected by edges. A maximal clique is a clique that cannot be extended by adding any nodes. In particular, the maximal clique with the most nodes is  the maximum clique of a graph.

\noindent\textbf{Searching for Maximal cliques.} To generate hypotheses, RANSAC-based methods repeatedly take random samples from the correspondence set. Nevertheless, they fail to fully mine the affinity relationships between correspondences. Theoretically, inliers would form cliques in the graph, because inliers are usually geometrically compatible with each other. Previous works~\cite{yang2020teaser, lin2022k,parra2019practical,lin2022planted} focus on searching for maximum cliques in the graph, however, the maximum clique is a very tight constraint that only focuses on the global consensus information in a graph. Instead, we loosen the constraint and leverage maximal cliques to mine more local graph information. 

By using the ${\emph igraph\_maximal\_cliques}$ function in the igraph\footnote{https://igraph.org} C++ library, which makes use of a modified Bron-Kerbosch algorithm~\cite{eppstein2010listing}, the search of maximal cliques can be very efficient. The process's worst time complexity is ${\mathcal{O}}(d(n-d)3^{(d/3)})$, where $d$ is the degeneracy of the graph. Note that $d$ is typically small in our problem because the graph is usually sparse when dealing with point cloud correspondences. 

\noindent\textbf{Node-guided Clique Selection.}
After executing the maximal clique searching procedure, we obtain the maximal clique set ${\bf\emph{MAC}}_{initial}$. In practice, ${\bf\emph{MAC}}_{initial}$ usually contains tens of thousands of maximal cliques, which will make it very time-consuming if we consider all maximal cliques. We introduce a node-guided clique selection method in this section to reduce $|{\bf\emph{MAC}}_{initial}|$. First, we calculate the weight for each clique in ${\bf\emph{MAC}}_{initial}$. Given a clique ${C_i} = ({{\bf V}_i},{{\bf E}_i})$, the weight ${w_{{C_i}}}$ is calculated as: 
\begin{equation}
    {w_{{C_i}}} = \sum\limits_{{e_j} \in {{\bf E}_i}} {{w_{{e_j}}}},
\end{equation} 
where ${w_{{e_j}}}$ represents the weight of edge ${e_j}$ in ${{\mathbf{W}}_{SOG}}$. A node may be included by multiple maximal cliques and we only retain the one with the greatest weight for that node. Then, duplicated cliques are removed from the rest, obtaining ${\bf\emph{MAC}}_{selected}$. The motivation behind this is to use information about the local geometric structure around graph nodes to find the best consistent set of corresponding nodes. 
It is clear that the number of maximal cliques $|{\bf\emph{MAC}}_{selected}|$ will not exceed $|{\bf V}|$. We could send these maximal cliques directly to the following stages for 3D registration. However, when $|{\bf V}|$ is quite large, the number of retained maximal cliques can still be very large. Here, we propose several techniques to further filter the maximal cliques.
\begin{itemize}
    \item {\textbf{Normal consistency.}} In the maximal cliques, we find that the normal consistency is satisfied between each correspondence. Given two correspondences ${{\mathbf c}_i}=({\mathbf p}_i^s,{\mathbf p}_i^t)$, ${{\mathbf c}_j}=({\mathbf p}_j^s,{\mathbf p}_j^t)$ and the normal vectors $\mathbf{n}_i^s,\mathbf{n}_j^s,\mathbf{n}_i^t,\mathbf{n}_j^t$ at the four points, the angular difference ${\alpha}_{ij}^s = \angle (\mathbf{n}_i^s,\mathbf{n}_j^s)$,
    ${\alpha}_{ij}^t = \angle (\mathbf{n}_i^t, \mathbf{n}_j^t)$ between the normal vectors can be calculated then. The following inequality ought to hold if ${\mathbf c}_i$ and ${\mathbf c}_j$ are normal consistent:
    \begin{equation}
        \left|{\rm sin}{{\alpha}_{ij}^s}-{\rm sin}{{\alpha}_{ij}^t}\right|<{t_\alpha},
    \end{equation}
    where $t_\alpha$ is a threshold for determining whether the angular differences are similar.
    \item {\textbf{Clique ranking.}} We organize ${\bf\emph{MAC}}_{selected}$ in a descending order using the clique's weight ${w_{{C_i}}}$. The top-$K$ ones are supposed to be more likely to produce correct hypotheses. This makes it flexible to control the number of hypotheses.
\end{itemize}

These techniques' experimental analysis is presented in  Sec.~\ref{sec:ablation}.
\subsection{Hypothesis Generation and Evaluation}
Each maximal clique filtered from the previous step represents a consistent set of correspondences. By applying the SVD algorithm to each consistency set, we can obtain a set of 6-DoF pose hypotheses. 
\begin{itemize}
    \item \textbf{Instance-equal SVD.} Transformation estimation of correspondences is often implemented with SVD. Instance-equal means that the weights of all correspondences are equal.
    \item \textbf{Weighted SVD.} 
    Assigning weights to correspondences is commonly adopted by recent PCR methods~\cite{chen2022sc2,qin2022geometric,choy2020deep,pais20203dregnet}. Correspondence weights can be derived by solving the eigenvectors of a compatibility matrix constructed for a compatibility graph. Here, we take the primary eigenvalues of ${{\mathbf{W}}_{SOG}}$ as correspondence weights.
\end{itemize}

The final goal of MAC is to estimate the optimal 6-DoF rigid transformation (composed of a rotation pose ${{\mathbf{R}}^*}\in SO(3)$ and a translation pose ${{\mathbf{t}}^*}\in {\mathbb{R}^3}$) that maximizes the objective function as follow:
\begin{equation}
    ({{\mathbf R}^*},{{\mathbf t}^*}) = \arg \mathop {\max }\limits_{{\mathbf R},{\mathbf t}} \sum\limits_{i = 1}^N {s({{\mathbf c}_i})},
\end{equation}
where ${{\mathbf c}_i} \in {\mathbf C}_{initial}$, $N=|{\mathbf C}_{initial}|$, and $s({{\mathbf c}_i})$ represents the score of ${{\mathbf c}_i}$.
We consider several RANSAC hypothesis evaluation metrics here ~\cite{yang2021toward}, including mean average error (MAE), mean square error (MSE) and inlier count. Their behaviors will be experimentally compared in Sec.~\ref{sec:ablation}.
The best hypothesis is taken to perform 3D registration then. 
\section{Experiments}
\label{sec:experiment}
\subsection{Experimental Setup}
\noindent\textbf{Datasets.} We consider four datasets, i.e, the object-scale dataset U3M~\cite{mian2006novel}, the scene-scale indoor datasets 3DMatch~\cite{zeng20173dmatch} \& 3DLoMatch~\cite{huang2021predator}, and the scene-scale outdoor dataset KITTI~\cite{geiger2012we}. U3M has 496 point cloud pairs. 3DLoMatch is the subset of 3DMatch, where the overlap rate of the point cloud pairs ranges from 10\% to 30\%, which is very challenging. For KITTI, we follow~\cite{bai2021pointdsc,chen2022sc2} and  obtain 555 pairs of point clouds for testing.

\noindent\textbf{Evaluation Criteria.} We follow~\cite{yang2021sac} that employs the root mean square error (RMSE) metric to evaluate the 3D point cloud registration performance on the U3M object-scale dataset. In addition, we employ the rotation error (RE) and translation error (TE) to evaluate the registration results on the scene-scale dataset. By referring to the settings in~\cite{choy2020deep},  the registration is considered successful when the RE $\leq$ 15\textdegree, TE $\leq$ 30 cm on 3DMatch \& 3DLoMatch datasets, and RE $\leq$ 5\textdegree, TE $\leq$ 60 cm on KITTI dataset. We define a dataset's registration accuracy as the ratio of success cases to the number of point cloud pairs to be registered.

\noindent\textbf{Implementation Details.} Our method is implemented in C++ based on the point cloud library (PCL)~\cite{rusu20113d} and igraph library. For U3M, we use the Harris3D (H3D)~\cite{sipiran2011harris} keypoint detector and the signatures of histograms of orientation (SHOT)~\cite{tombari2010unique} descriptor for initial correspondence generation as in ~\cite{yang2019ranking}. For 3DMatch and 3DLoMatch datasets, we use the fast point features histograms (FPFH)~\cite{rusu2009fast} descriptor and fully convolutional geometric features (FCGF)~\cite{choy2019fully} descriptor to generate the initial correspondence set. The main steps in the comparative experimental sections are SOG construction, searching node-guided maximal cliques, hypotheses generation by instance-equal SVD and evaluation by MAE. Default values for compatibility threshold $t_{cmp}$ and distance parameter ${d_{cmp}}$ mentioned in Sec.~\ref{subsec:graph} are 0.99 and 10 pr respectively; if input matches exceed 5000, ${t_{cmp}}$ is set to 0.999 to reduce computation. Here, `pr' is a distance unit called point cloud resolution~\cite{yang2019ranking}. Normal vectors are calculated using the NormalEstmation class of PCL with the 20 nearest neighboring points. When searching maximal cliques, the lower bound on clique size is set to 3 with no upper bound defined. All experiments were implemented with an Intel 12700H CPU and 32 GB RAM. 

\begin{figure}[t]
\centering
\includegraphics[width=0.9\linewidth]{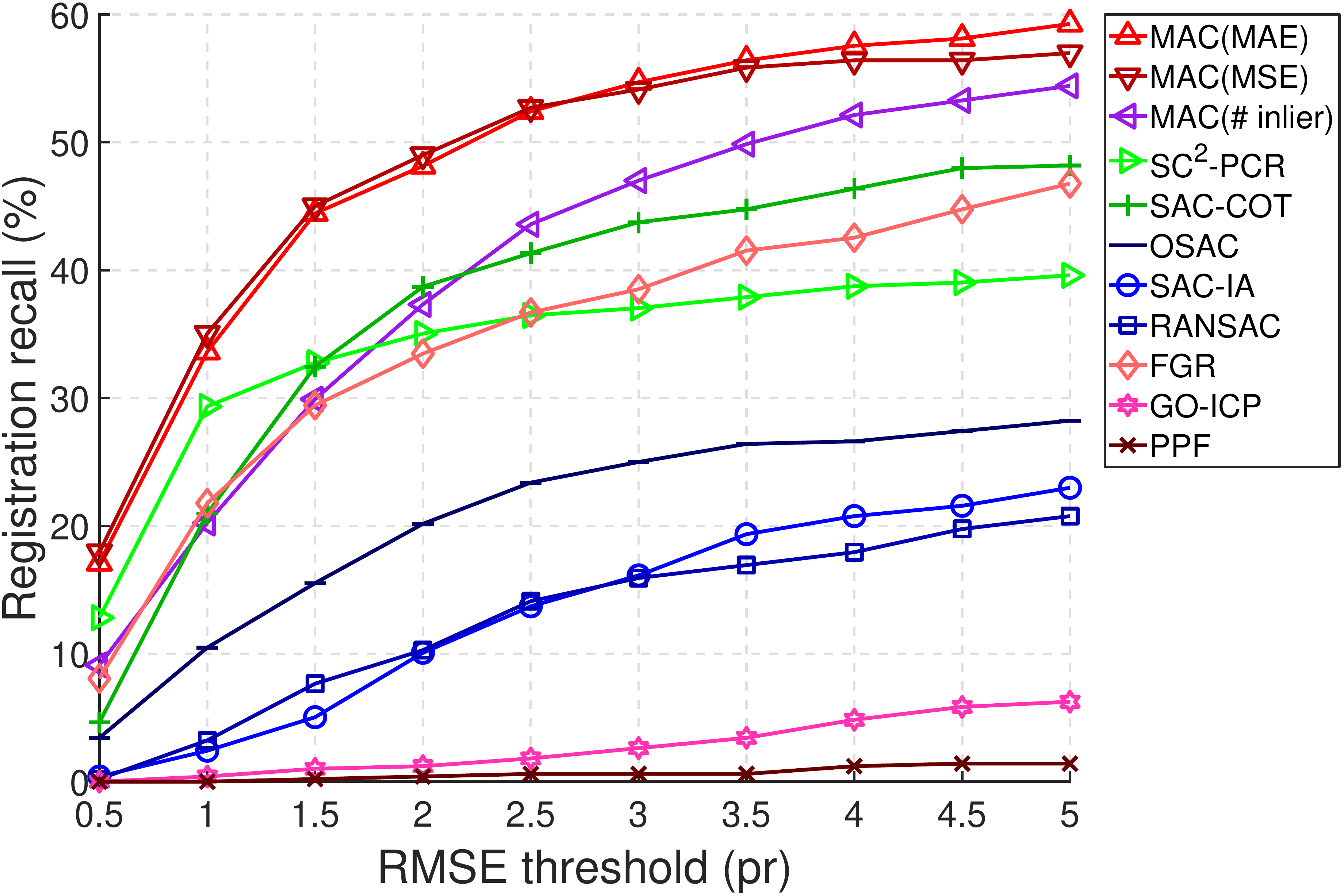}
\caption{Registration performance of tested point cloud registration methods on U3M.}
\label{fig:U3M_variants}
\end{figure}

\subsection{Results on U3M Dataset} 
 We perform an extensive comparison in Fig.~\ref{fig:U3M_variants}. Here, the following methods are tested, including SAC-COT\cite{yang2021sac}, OSAC\cite{yang2016fast}, SAC-IA\cite{rusu2009fast}, RANSAC\cite{fischler1981random}, $\rm{SC}^2$-PCR\cite{chen2022sc2}, FGR\cite{zhou2016fast}, GO-ICP\cite{yang2015go}, and PPF\cite{drost2010model}, where the former four are RANSAC-based methods. The RMSE threshold is varied from 0.5 pr to 5 pr with a step of 0.5 pr. 
 
 The results indicate that MAC performs best and significantly outperforms all tested RANSAC-fashion estimators, such as SAC-COT, OSAC, SAC-IA, and RANSAC. The registration performance of MAC based on the MAE evaluation metric is the best on U3M.
 
\subsection{Results on 3DMatch \& 3DLoMatch Datasets}
\begin{table}[t]
  \centering
  \resizebox{\linewidth}{!}{
    \begin{tabular}{l|rrr|rrr}
    \hline
    \multirow{2}[2]{*}{} & \multicolumn{3}{c|}{FPFH} & \multicolumn{3}{c}{FCGF} \\
          & \multicolumn{1}{l}{RR(\%)} & \multicolumn{1}{l}{RE(\textdegree)} & \multicolumn{1}{l|}{TE(cm)} & \multicolumn{1}{l}{RR(\%)} & \multicolumn{1}{l}{RE(\textdegree)} & \multicolumn{1}{l}{TE(cm)} \\
    \hline
    \emph{i) Traditional} & & & & & & \\
    SM\cite{leordeanu2005spectral} & 55.88 & 2.94 & 8.15 & 86.57 & 2.29 & 7.07 \\
    FGR\cite{zhou2016fast} & 40.91 & 4.96 & 10.25 & 78.93 & 2.90 & 8.41 \\
    RANSAC-1M\cite{fischler1981random} & 64.20 & 4.05 & 11.35 & 88.42 & 3.05 & 9.42 \\
    RANSAC-4M\cite{fischler1981random} & 66.10 & 3.95 & 11.03 & 91.44 & 2.69 & 8.38 \\
    GC-RANSAC\cite{barath2018graph} & 67.65 & 2.33 & 6.87 & 92.05 & 2.33 & 7.11 \\
    TEASER++\cite{yang2020teaser} & 75.48 & 2.48 & 7.31 & 85.77 & 2.73 & 8.66 \\
    CG-SAC\cite{quan2020compatibility} & 78.00 & 2.40 & 6.89 & 87.52 & 2.42 & 7.66 \\
    $\rm{SC}^2$-PCR\cite{chen2022sc2} & \underline{83.73}  & \underline{2.18}  & {6.70} & \underline{93.16}  & \underline{2.09}  & \underline{6.51} \\
    \hline
    \emph{ii) Deep learned} & & & & & & \\
    3DRegNet\cite{pais20203dregnet} & 26.31 & 3.75 & 9.60 & 77.76 & 2.74 & 8.13 \\
    DGR\cite{choy2020deep}   &  32.84  & 2.45 & 7.53  &  88.85 & 2.28 & 7.02 \\
    DHVR\cite{lee2021deep}  & 67.10 & 2.78 & 7.84 & 91.93 & 2.25 & 7.08 \\
    PointDSC\cite{bai2021pointdsc} & 72.95 & \underline{2.18}  & \underline{6.45}  &  91.87 & {2.10}  & {6.54} \\
    \hline
    MAC & \textbf{84.10}  & \textbf{1.96}  & \textbf{6.18}  &  \textbf{93.72} &  \textbf{1.89}  & \textbf{6.03} \\
    \hline
    \end{tabular}}
    \caption{Registration results on 3DMatch dataset.}
  \label{tab:3dmatch5000}%
\end{table}%

\begin{table}[t]
  \centering
  \resizebox{\linewidth}{!}{
    \begin{tabular}{l|rrr|rrr}
    \hline
    \multirow{2}[2]{*}{} & \multicolumn{3}{c|}{FPFH} & \multicolumn{3}{c}{FCGF} \\
          & \multicolumn{1}{l}{RR(\%)} & \multicolumn{1}{l}{RE(\textdegree)} & \multicolumn{1}{l|}{TE(cm)} & \multicolumn{1}{l}{RR(\%)} & \multicolumn{1}{l}{RE(\textdegree)} & \multicolumn{1}{l}{TE(cm)} \\
    \hline
    \emph{i) Traditional} & & & & & & \\
    RANSAC-1M\cite{fischler1981random} &0.67 &10.27 &15.06 & 9.77& 7.01 & 14.87\\
    RANSAC-4M\cite{fischler1981random} &0.45 &10.39 & 20.03&10.44 & 6.91 & 15.14\\
    TEASER++\cite{yang2020teaser}  & 35.15 & 4.38  & 10.96  & 46.76  &  4.12 &  12.89\\
    $\rm{SC}^2$-PCR\cite{chen2022sc2} &  \underline{38.57} & \underline{4.03}  & 10.31  & \underline{58.73}  & \underline{3.80}  & \underline{10.44} \\
    \hline
    \emph{ii) Deep learned} & & & & & & \\
    DGR\cite{choy2020deep}   &  19.88  & 5.07 & 13.53  & 43.80  & 4.17 & 10.82 \\
    PointDSC\cite{bai2021pointdsc} & 20.38  & 4.04  & \underline{10.25}  & 56.20  & 3.87  & {10.48} \\
    \hline
    MAC & \textbf{40.88} & \textbf{3.66} & \textbf{9.45} & \textbf{59.85} & \textbf{3.50} & \textbf{9.75}\\
    \hline
    \end{tabular}}
    \caption{Registration results on 3DLoMatch dataset.}
  \label{tab:3dlomatch5000}%
\end{table}%

\noindent\textbf{PCR methods comparison.}
Both geometric-only and deep-learned methods are considered for comparison, including
SM\cite{leordeanu2005spectral}, FGR\cite{zhou2016fast}, RANSAC\cite{fischler1981random}, TEASER++\cite{yang2020teaser}, CG-SAC\cite{quan2020compatibility}, $\rm{SC}^2$-PCR\cite{chen2022sc2}, 3DRegNet\cite{pais20203dregnet}, DGR\cite{choy2020deep}, DHVR\cite{lee2021deep} and PointDSC\cite{bai2021pointdsc}.
 Results are shown in Tables~\ref{tab:3dmatch5000} and~\ref{tab:3dlomatch5000}. 
 
 The following conclusions can be made: 1) regardless of which descriptor is used, MAC outperforms all compared methods on both 3DMatch and 3DLoMatch datasets, indicating its strong ability to register indoor scene point clouds; 2) even compared with deep-learned methods, MAC still achieves better performance without any data training; 3) in addition to the registration recall (RR) metric, MAC achieves the best RE and TE metrics. This indicates that registrations by MAC are very accurate and MAC is able to align low overlapping data.
\begin{table}[t]
\centering
\resizebox{\linewidth}{!}{
\begin{tabular}{l|lllll|lllll}
\hline
\multirow{2}{*}{\# Samples} & \multicolumn{5}{c|}{3DMatch RR(\%)}     & \multicolumn{5}{c}{3DLoMatch RR(\%)}    \\
                            & 5000 & 2500 & 1000 & 500  & 250  & 5000 & 2500 & 1000 & 500  & 250  \\ \hline
FCGF\cite{choy2019fully}                  & 85.1 & 84.7 & 83.3 & 81.6 & 71.4 & 40.1 & 41.7 & 38.2 & 35.4 & 26.8 \\
SpinNet\cite{ao2021spinnet}
& 88.6 & 86.6 & 85.5 & 83.5 & 70.2 & 59.8 & 54.9 & 48.3 & 39.8 & 26.8 \\
Predator\cite{huang2021predator}                 & 89.0 & 89.9 & 90.6 & 88.5 & 86.6 & 59.8 & 61.2 & 62.4 & 60.8 & 58.1 \\
CoFiNet\cite{yu2021cofinet}                     
& 89.3 & 88.9 & 88.4 & 87.4 & 87.0 & 67.5 & 66.2 & 64.2 & 63.1 & 61.0 \\
GeoTransformer\cite{qin2022geometric}
& 92.0 & 91.8 & 91.8 & 91.4 & 91.2 & 75.0 & 74.8 & 74.2 & 74.1 & 73.5 \\ \hline
      \multirow{2}{*}{FCGF+MAC}      &  91.3    &    92.2  &  91.6    &  90.4    &   85.6   &   57.2   & 56.0   & 52.6    &  42.4    &  32.1    \\
               &   6.2$\uparrow$   &  7.5$\uparrow$    & 8.3$\uparrow$     &   8.8$\uparrow$   & 14.2$\uparrow$     &  17.1$\uparrow$    & 14.3$\uparrow$     &  14.4$\uparrow$    &  7.0$\uparrow$    &  5.3$\uparrow$    \\
    \hdashline
     \multirow{2}{*}{SpinNet+MAC}   &   95.3   & 95.1  &  93.3  &   91.4 & 81.2   &  72.8  & 69.9  &  59.2  &   54.8   & 32.1    \\
              &  6.7$\uparrow$   &  8.5$\uparrow$    &  7.8$\uparrow$    &  7.9$\uparrow$    &   11.0$\uparrow$   &  13.0$\uparrow$    &  15.0$\uparrow$    &  10.9$\uparrow$    &  15.0$\uparrow$    &  5.3$\uparrow$    \\
    \hdashline
    \multirow{2}{*}{Predator+MAC}        &  94.6    &  94.4    &   94.0   &  93.5    &  92.3    &  70.9    &   70.4   &   69.8   &   67.2   &   64.1   \\
                 &   5.6$\uparrow$   &  4.5$\uparrow$    & 3.4$\uparrow$     &  5.0$\uparrow$    &   5.7$\uparrow$   &  11.1$\uparrow$    & 9.2$\uparrow$     &  7.4$\uparrow$  &   6.4$\uparrow$   &  6.0$\uparrow$    \\
    \hdashline
       \multirow{2}{*}{CoFiNet+MAC}    &   94.1   &  94.4    &   94.5   &  93.8    &   92.7   &  71.6  &   71.5   &   70.6   &   69.2    &   68.1   \\
                &    4.8$\uparrow$  &   5.5$\uparrow$   &   6.1$\uparrow$   &  6.4$\uparrow$  &   5.7$\uparrow$   &   4.1$\uparrow$   &  5.3$\uparrow$  &   6.4$\uparrow$   &   6.1$\uparrow$  &   7.1$\uparrow$  \\
    \hdashline
      \multirow{2}{*}{GeoTransformer+MAC}    &   \bf{95.7}   &   \bf{95.7}   &   \bf{95.2}   &    \bf{95.3}  &   \bf{94.6}   &  \bf{78.9}    &   \bf{78.7}   &   \bf{78.2}   &   \bf{77.7}   &   \bf{76.6}   \\ 
               &   3.7$\uparrow$   &   3.9$\uparrow$   &  3.4$\uparrow$  &   3.9$\uparrow$   &   3.4$\uparrow$   &  3.9$\uparrow$    &   3.9$\uparrow$   &  4.0$\uparrow$  & 3.6$\uparrow$ &   3.1$\uparrow$   \\
               \hline
\end{tabular}}
\caption{Performance boosting for deep-learned methods when combined with MAC.}
\label{tab:desc}
\end{table}

\noindent\textbf{Boosting deep-learned methods with MAC.} Several kinds of state-of-the-art deep-learned methods are integrated with MAC for evaluation. The considered methods are FCGF\cite{choy2019fully}, 
SpinNet\cite{ao2021spinnet}, Predator\cite{huang2021predator}, CoFiNet\cite{yu2021cofinet} and GeoTransformer\cite{qin2022geometric}. Each method is tested under a different number of samples, which refer to the number of sampled points or correspondences. Results are reported in Table~\ref{tab:desc}.

Remarkably, MAC dramatically improves the registration recall under all tested methods on both 3DMatch and 3DLoMatch datasets. Notably, the performance of SpinNet, Predator and CoFiNet after boosting by MAC exceeds that of GeoTransformer. MAC working with GeoTransformer achieves state-of-the-art registration recall of {\bf{95.7\% / 78.9\%}} on 3DMatch / 3DLoMatch. The results suggest that: 1) MAC can greatly boost existing deep-learned methods; 2) MAC is not sensitive to the number of samples.

\subsection{Results on KITTI Dataset}
\begin{table}[t]
  \centering
  \resizebox{\linewidth}{!}{
    \begin{tabular}{l|rrr|rrr}
    \hline
    \multirow{2}[2]{*}{} & \multicolumn{3}{c|}{FPFH} & \multicolumn{3}{c}{FCGF} \\
          & \multicolumn{1}{l}{RR(\%)} & \multicolumn{1}{l}{RE(\textdegree)} & \multicolumn{1}{l|}{TE(cm)} & \multicolumn{1}{l}{RR(\%)} & \multicolumn{1}{l}{RE(\textdegree)} & \multicolumn{1}{l}{TE(cm)} \\
    \hline
    \emph{i) Traditional} & & & & & & \\
    FGR\cite{zhou2016fast}  & 5.23 & 0.86 & 43.84 & 89.54 & 0.46 & 25.72 \\
    TEASER++\cite{yang2020teaser}   &  91.17  &  1.03  & 17.98 & 94.96   &  0.38  & \textbf{13.69} \\
    RANSAC\cite{fischler1981random}  &  74.41  &  1.55  &  30.20 & 80.36  &  0.73  & 26.79 \\
    CG-SAC\cite{quan2020compatibility}  & 74.23  & 0.73  & 14.02 & 83.24  & 0.56  & 22.96 \\
    $\rm{SC}^2$-PCR\cite{chen2022sc2} & \underline{99.28}  &  \underline{0.39}  & {8.68}  & \textbf{97.84}  & \textbf{0.33}  & 20.58 \\
    \hline
    \emph{ii) Deep learned} & & & & & & \\
    DGR\cite{choy2020deep}   &   77.12   &  1.64  & 33.10   &  96.90 & \underline{0.34}  & 21.70 \\
    PointDSC\cite{bai2021pointdsc} &  \underline{98.92}  &  \textbf{0.38}  & \textbf{8.35}  &  \textbf{97.84} &  \textbf{0.33}  & {20.32} \\
    \hline
    MAC &  \textbf{99.46} & 0.40  & \underline{8.46}  &  \textbf{97.84} & \underline{0.34} & \underline{19.34} \\
    \hline
    \end{tabular}}
    \caption{Registration results on KITTI dataset.}
  \label{tab:kitti}%
\end{table}%
In Table~\ref{tab:kitti}, the results of DGR\cite{choy2020deep}, PointDSC\cite{bai2021pointdsc}, TEASER++\cite{yang2020teaser}, RANSAC\cite{fischler1981random}, CG-SAC\cite{quan2020compatibility}, $\rm{SC}^2$-PCR\cite{chen2022sc2} and MAC are reported for comparison. 
    
    As shown by the table, in terms of the registration recall performance, MAC presents the best and is tied for the best results with FPFH and FCGF descriptor settings, respectively. MAC also has a lower TE than the state-of-the-art geometric-only method $\rm{SC}^2$-PCR. Note that outdoor point clouds are significantly sparse and non-uniformly distributed. The registration experiments on the object, indoor scene, and outdoor scene datasets consistently verify that MAC holds good generalization ability  in different application contexts.

\subsection{Analysis Experiments}\label{sec:ablation}
\begin{table*}[t]
\resizebox{\linewidth}{!}{
\begin{tabular}{cc|cccccccccccc|clc|clc|clc}
\hline
\multicolumn{2}{c|}{}                & FOG        & SOG        & GC         & MC         & NG         & NC         & CR         & SVD        & W-SVD      & MAE        & MSE        & \# inlier  & \multicolumn{3}{c|}{RR(\%)}                      & \multicolumn{3}{c|}{RE(\textdegree)}           & \multicolumn{3}{c}{TE(cm)}                     \\ \hline
\multirow{11}{*}{FPFH} & 1)          &            & \checkmark &            &            & \checkmark &            &            & \checkmark &            &            &            & \checkmark & 83.86          & / & 39.14          & 2.17          & / & 4.01          & 6.51          & / & 9.94          \\
                       & 2)          &            & \checkmark & \checkmark &            & \checkmark &            &            & \checkmark &            &            &            & \checkmark & 77.02          & / & 26.61          & 2.10          & / & 3.83          & 6.19          & / & 9.49          \\
                       & 3)          & \checkmark &            &            &            & \checkmark &            &            & \checkmark &            &            &            & \checkmark & 82.26          & / & 39.02          & 2.12          & / & 3.98          & 6.43          & / & 9.89          \\
                       & 4)          &            & \checkmark &            &            &            &            &            & \checkmark &            &            &            & \checkmark & 83.49          & / & 38.91          & 2.22          & / & 4.11          & 6.65          & / & 10.05         \\
                       & 5)          &            & \checkmark &            &            & \checkmark &            &            &            & \checkmark &            &            & \checkmark & 83.67          & / & 38.85          & 2.15          & / & 4.03          & 6.53          & / & 9.82          \\
                       & 6)          &            & \checkmark &            &            & \checkmark &            &            & \checkmark &            & \checkmark &            &            & \textbf{84.10} & / & \textbf{40.88} & \textbf{1.96} & / & \textbf{3.66} & \textbf{6.18} & / & \textbf{9.45} \\
                       & 7)          &            & \checkmark &            &            & \checkmark &            &            & \checkmark &            &            & \checkmark &            & 82.93          & / & 39.98          & 1.95          & / & 3.66          & 6.12          & / & 9.48          \\
                       & 8)          &            & \checkmark &            &            & \checkmark & \checkmark &            & \checkmark &            &            &            & \checkmark & 82.44          & / & 38.46          & 2.16          & / & 3.97          & 6.41          & / & 9.85          \\
                       & 9)          &            & \checkmark &            & \checkmark & \checkmark &            &            & \checkmark &            &            &            & \checkmark & 74.06          & / & 31.11          & 2.08          & / & 3.89          & 6.17          & / & 9.82          \\
                       & 10) Top100  &            &     \checkmark       &            &            & \checkmark &            & \checkmark & \checkmark &            &            &            & \checkmark & 82.01          & / & 37.79          & 2.13          & / & 4.02          & 6.42          & / & 9.82          \\
                       & 11) Top200  &            &    \checkmark        &            &            & \checkmark &            & \checkmark & \checkmark &            &            &            & \checkmark & 83.18          & / & 38.85          & 2.16          & / & 4.08          & 6.55          & / & 9.91          \\
                       & 12) Top500  &            &   \checkmark         &            &            & \checkmark &            & \checkmark & \checkmark &            &            &            & \checkmark & 83.06          & / & 38.85          & 2.14          & / & 4.03          & 6.47          & / & 9.81          \\
                       & 13) Top1000 &            &   \checkmark         &            &            & \checkmark &            & \checkmark & \checkmark &            &            &            & \checkmark & 83.30          & / & 38.91          & 2.16          & / & 4.05          & 6.53          & / & 9.84          \\
                       & 14) Top2000 &            &   \checkmark         &            &            & \checkmark &            & \checkmark & \checkmark &            &            &            & \checkmark & 83.36          & / & 38.79          & 2.14          & / & 4.02          & 6.52          & / & 9.78          \\ \hline
\multirow{11}{*}{FCGF} & 1)          &            & \checkmark &            &            & \checkmark &            &            & \checkmark &            &            &            & \checkmark & 93.41          & / & 59.80          & 2.04          & / & 3.78          & 6.33          & / & 10.16         \\
                       & 2)          &            & \checkmark & \checkmark &            & \checkmark &            &            & \checkmark &            &            &            & \checkmark & 91.68          & / & 49.97          & 1.99          & / & 3.64          & 6.23          & / & 9.90          \\
                       & 3)          & \checkmark &            &            &            & \checkmark &            &            & \checkmark &            &            &            & \checkmark & 93.35          & / & 59.24          & 2.04          & / & 3.67          & 6.28          & / & 9.99          \\
                       & 4)          &            & \checkmark &            &            &            &            &            & \checkmark &            &            &            & \checkmark & 92.91          & / & 59.07          & 2.06          & / & 3.88          & 6.33          & / & 10.20         \\
                       & 5)          &            & \checkmark &            &            & \checkmark &            &            &            & \checkmark &            &            & \checkmark & 93.16          & / & 59.46          & 2.04          & / & 3.76          & 6.26          & / & 10.00         \\
                       & 6)          &            & \checkmark &            &            & \checkmark &            &            & \checkmark &            & \checkmark &            &            & \textbf{93.72} & / & \textbf{59.85} & \textbf{1.89} & / & \textbf{3.50} & \textbf{6.03} & / & \textbf{9.75} \\
                       & 7)          &            & \checkmark &            &            & \checkmark &            &            & \checkmark &            &            & \checkmark &            & 93.59          & / & 59.01          & 1.86          & / & 3.49          & 6.00          & / & 9.61          \\
                       & 8)          &            & \checkmark &            &            & \checkmark & \checkmark &            & \checkmark &            &            &            & \checkmark & 93.28          & / & 59.63          & 2.02          & / & 3.73          & 6.24          & / & 9.98          \\
                       & 9)          &            & \checkmark &            & \checkmark & \checkmark &            &            & \checkmark &            &            &            & \checkmark & 87.86          & / & 49.35          & 2.00          & / & 3.61          & 6.09          & / & 9.60          \\
                       & 10) Top100  &            &    \checkmark        &            &            & \checkmark &            & \checkmark & \checkmark &            &            &            & \checkmark & 92.42          & / & 57.44          & 2.00          & / & 3.75          & 6.21          & / & 10.00         \\
                       & 11) Top200  &            &   \checkmark         &            &            & \checkmark &            & \checkmark & \checkmark &            &            &            & \checkmark & 93.22          & / & 57.83          & 2.01          & / & 3.75          & 6.29          & / & 10.06         \\
                       & 12) Top500  &            &  \checkmark          &            &            & \checkmark &            & \checkmark & \checkmark &            &            &            & \checkmark & 93.22          & / & 58.90          & 2.02          & / & 3.78          & 6.33          & / & 10.02         \\
                       & 13) Top1000 &            &   \checkmark         &            &            & \checkmark &            & \checkmark & \checkmark &            &            &            & \checkmark & 93.35          & / & 59.40          & 2.05          & / & 3.78          & 6.32          & / & 10.18         \\
                       & 14) Top2000 &            &   \checkmark         &            &            & \checkmark &            & \checkmark & \checkmark &            &            &            & \checkmark & 93.35          & / & 59.52          & 2.04          & / & 3.78          & 6.33          & / & 10.19         \\ \hline
\end{tabular}}
\caption{Analysis experiments on 3DMatch / 3DLoMatch. \textbf{FOG}: First order compatibility graph. \textbf{SOG}: Second order compatibility graph. \textbf{GC}: Use geometric consistency to preliminarily perform outlier rejection. \textbf{MC}: Search the {\bf{\emph{maximum}}} clique instead of {\bf{\emph{maximal}}} cliques. \textbf{NG}: Node-guided clique selection. \textbf{NC}: Normal consistency. \textbf{CR}: Clique ranking. \textbf{W-SVD}: Weighted SVD.}
\label{tab:ablation1}
\end{table*}

In this section, we perform ablation studies and analysis experiments on both 3DMatch and 3DLoMatch datasets. We progressively experiment with the techniques proposed in Sec.~\ref{sec:method}, and the results are shown in Table~\ref{tab:ablation1}. The quality of generated hypotheses is analyzed in Table~\ref{tab:truehypo}. The performance upper bound is studied in Table~\ref{tab:evaluation_difficulty}. Table~\ref{tab:time_consumption} presents the time efficiency analysis of MAC.

\noindent\textbf{Performing feature matching selection.} Before 3D registration, a popular way is to perform outlier rejection to reduce the correspondence set. Here we employ 
 geometric consistency (GC)~\cite{chen20073d}, which is independent of the feature space and associates the largest consistent cluster relating to the compatibility among correspondences. 
 
 By comparing Row 1 and 2 of Table~\ref{tab:ablation1}, GC has a negative impact on MAC performance, potentially due to that some inliers are also removed in this process. This demonstrates that MAC can still perform well even if the initial correspondence set is directly utilized as input without any filtering.

\noindent\textbf{Graph construction choices.} We test the performance of MAC by using different graph construction approaches. 

As shown in Row 1 and 3 of Table~\ref{tab:ablation1}, the registration recall obtained by using SOG is 1.6\% higher than using FOG when combined with FPFH, and 0.06\% higher when combined with FCGF on 3DMatch. Also, the registration recall obtained by using SOG is 0.12\% higher than using FOG when combined with FPFH, and 0.56\% higher when combined with FCGF on 3DLoMatch. Therefore, SOG is more suitable for MAC. Detailed analyzing descriptions can be found in the supplementary. 

\noindent\textbf{Maximum or maximal clique.} To justify the advantages of maximal cliques, we change the search strategy of MAC to the maximum cliques and test the registration performance. 

As shown in Row 1 and 9 in Table~\ref{tab:ablation1}, applying maximal cliques surpasses maximum by 9.8\% when combined with FPFH, and 5.55\% higher when combined with FCGF on 3DMatch. Besides, the registration recall obtained by using maximal cliques is 8.03\% higher than using the maximum cliques when combined with FPFH and 10.45\% higher when combined with FCGF on 3DLoMatch. There are several reasons for this: 1) maximal cliques include the maximum cliques and additionally consider local graph constraints, so the search for maximal cliques can make use of both local and global information in the compatibility graph; 2) the maximum clique is a very tight constraint which requires maximizing the number of mutually compatible correspondences, but it does not guarantee the optimal result. 

\noindent\textbf{Node-guided clique selection.} We compare the performance with and without node-guided (NG) clique selection for maximal cliques search. 

Comparing Row 1 and 4 in Table~\ref{tab:ablation1}, using NG achieves a recall improvement of 0.37\% when combined with FPFH, and 0.5\% improvement when combined with FCGF on 3DMatch. Also, using NG achieves a recall improvement of 0.23\% with FPFH and 0.73\% improvement with FCGF on 3DLoMatch. It is worth noting that while NG improves recall, the mean RE and mean TE are also decreasing. For example, NG reduces the mean RE by 0.1\textdegree and the mean TE by 0.11 cm with FPFH on 3DLoMatch. NG effectively reduces the number of calculations in the subsequent steps and promises accurate hypotheses.

\noindent\textbf{Different approaches for clique filtering.} We test the effectiveness of the two filtering methods, normal consistency and clique ranking.

\emph{1) Normal consistency:} comparing Row 1 and 8 in Table~\ref{tab:ablation1}, NC slightly degrades MAC's performance. \emph{2) Clique ranking:} Row 10 to 14 demonstrate that the registration recall tends to increase as $K$ increases, suggesting that larger $K$ yields a subset of cliques that generate more correct hypotheses. Remarkably, setting $K$ to 100 can already achieve outstanding performance. 

\noindent\textbf{Employing instance-equal or weighted SVD.} The comparisons of instance-equal and weighted SVD are shown in Rows 1 and 5 of Table~\ref{tab:ablation1}. 

Weighted SVD is slightly inferior to instance-equal SVD. This suggests that 
samples in MACs are already very consistent, indicating no additional weighting strategies are required.

\noindent\textbf{Varying hypothesis evaluation metrics.} Here we compare three evaluation metrics, including MAE, MSE and inlier count, for MAC hypothesis evaluation. 

As shown in Row 1, 6 and 7, MAC with MAE achieves the best performance. In Table~\ref{tab:ablation1}, MAE achieves a recall improvement of 0.24\% when combined with FPFH, and 0.31\% improvement when combined with FCGF on 3DMatch compared with the commonly used inlier count metric. Also, MAE has a 1.74\% improvement when combined with FPFH, and 0.05\% when combined with FCGF on 3DLoMatch compared with inlier count. MAE is also very effective in reducing RE and TE. For instance, MAE reduces the mean RE by 0.35\textdegree and the mean TE by 0.49 cm with FPFH on 3DLoMatch.
\begin{table}[t]
\centering
\resizebox{\linewidth}{!}{
\begin{tabular}{c|cccc|cccc}
\hline
\multirow{3}{*}{} & \multicolumn{4}{c|}{3DMatch}    & \multicolumn{4}{c}{3DLoMatch} \\ \cline{2-9} 
   \# hypotheses   & \multicolumn{2}{c|}{RANSAC}      & \multicolumn{2}{c|}{MAC} & \multicolumn{2}{c|}{RANSAC}      & \multicolumn{2}{c}{MAC} \\ 
                  & FCGF & \multicolumn{1}{c|}{FPFH} & FCGF        & FPFH       & FCGF & \multicolumn{1}{c|}{FPFH} & FCGF       & FPFH       \\ \hline
100               &  10.45 & \multicolumn{1}{c|}{0.76}  & 61.94 & 50.67 & 1.25 &  \multicolumn{1}{c|}{0.05} & 30.47 & 12.22 \\
200               &  20.76 & \multicolumn{1}{c|}{1.50} & 119.20 & 89.27 & 2.52 & \multicolumn{1}{c|}{0.09} & 55.57 & 17.59 \\
500               & 51.74 & \multicolumn{1}{c|}{3.68} & 269.06 & 162.41 & 6.21 & \multicolumn{1}{c|}{0.21} & 109.32 & 23.32 \\
1000              & 103.65 & \multicolumn{1}{c|}{7.39} & 456.18 & 217.32 & 12.43 & \multicolumn{1}{c|}{0.41} & 156.11 & 26.02 \\
2000              & 208.24 & \multicolumn{1}{c|}{14.90} & 669.32 & 254.13 & 24.80 & \multicolumn{1}{c|}{0.81} & 202.12 & 29.31 \\ \hline
\end{tabular}}
\caption{Comparison of the number of correct hypotheses generated by MAC and RANSAC on 3DMatch and 3DLoMatch.}
\label{tab:truehypo}
\end{table}
\begin{table}[t]
\centering
\resizebox{0.5\linewidth}{!}{
\begin{tabular}{c|cc}
\hline
       & \multicolumn{1}{c}{\begin{tabular}[c]{@{}c@{}}3DMatch\\ RR(\%)\end{tabular}} & \multicolumn{1}{c}{\begin{tabular}[c]{@{}c@{}}3DLoMatch\\ RR(\%)\end{tabular}} \\ \hline
MAC-1  &  98.46 &  91.24 \\
MAC-5  &  97.10 &  83.32    \\
MAC-10 &  96.43 &  77.93    \\
MAC-20 &  94.70 &  70.47    \\
MAC-50 &  91.13 &  56.37   \\ 
MAC-origin & 93.72 & 59.85 \\ \hline
\end{tabular}}
\caption{Registration recall on 3DMatch with FCGF setting based on judging MAC's hypotheses. MAC-$n$: a point cloud pair is considered alignable if at least $n$ hypotheses are correct.}
\label{tab:evaluation_difficulty}
\end{table}

\noindent\textbf{Comparison with RANSAC hypotheses.} 
We evaluate the quality of the generated hypotheses by comparing the hypotheses from RANSAC and MAC with the ground truth transformation. The results are shown in Table~\ref{tab:truehypo}. 

Compared to RANSAC, which randomly selects correspondences and generates hypotheses from the correspondence set without geometric constraints, MAC effectively generates more convincing hypotheses from maximal cliques in the compatibility graph, which fully exploits the consensus information in the graph. 

\noindent\textbf{The performance upper bound of MAC.} Given an ideal hypothesis evaluation metric, allowing a point cloud pair can be aligned as long as correct hypotheses can be generated. This can test the performance upper bound of MAC. We vary the judging threshold for the number of generated correct hypotheses and report the results in Table~\ref{tab:evaluation_difficulty}.

Impressively, MAC-1 achieves registration recalls of {\bf{98.46\% / 91.24\%}} on 3DMatch / 3DLoMatch. This indicates that even on low overlapping datasets, MAC is able to produce correct hypotheses for most point cloud pairs. In addition, we can deduce that MAC's performance can be further improved with better hypothesis evaluation metrics.

\noindent\textbf{Time consumption of MAC.}
\begin{table}[t]
\resizebox{\linewidth}{!}{
\begin{tabular}{c|cccc|c}
\hline
     \# correspondences & \begin{tabular}[c]{@{}c@{}}Graph \\ Construction\end{tabular} & \begin{tabular}[c]{@{}c@{}}Search\\ Maximal\\ Cliques\end{tabular} & \begin{tabular}[c]{@{}c@{}}Node-guided\\ Clique\\ Selection\end{tabular} & \begin{tabular}[c]{@{}c@{}}Pose\\ Estimation\end{tabular} & Total\\ 
     \hline
250  &    1.03 (14.55\%)   &   5.24 (74.01\%)  & 0.58 (8.19\%) &    0.23 (3.25\%)        &    7.08          \\
500  &     4.07 (17.54\%) &  15.67 (67.51\%) &  3.12 (13.44\%) &  0.35 (1.51\%)   &  23.21 \\
1000 &   16.90 (29.85\%) & 36.60 (64.65\%) & 1.88 (3.32\%) &   1.23 (2.18\%) &  56.61  \\
2500 &    153.92 (53.29\%)  &  104.03 (36.02\%) & 4.97 (1.72\%) & 25.93 (8.97\%) &  288.85               \\
5000 &    887.03 (27.16\%)  &  1579.61 (48.37\%)  &  65.40 (2.00\%)  &  733.38 (22.47\%) &  3265.42 \\ 
\hline
\end{tabular}}
\caption{Average consumed time (ms) per point cloud pair on the 3DMatch dataset. Predator is used for generating correspondences.}
\label{tab:time_consumption}
\end{table}
We employ Predator~\cite{huang2021predator} to generate correspondences with different magnitudes to test the time performance of MAC. The time consumption is reported in Table~\ref{tab:time_consumption}. 

The following observations can be made. 1) In general, MAC can complete 3D registration in only tens of milliseconds when the number of correspondences is smaller than 1000. Even with an input with 2500 correspondences, the time consumption is about 0.29 seconds. Note that MAC is implemented on the CPU only. 2) As the number of correspondences increases from 250 to 2500, there is an increase in time cost for graph construction due to ${\mathbf{W}}_{SOG}$ computation taking more time. 3) When the number of correspondences reaches 5000, there is a large rise in the time cost of MAC's registration. The significant increase in the input size makes the search for maximal cliques more time-consuming. However, MAC is not sensitive to the cardinality of the input correspondence set, as verified in Table~\ref{tab:desc}. Hence, using sparse inputs for MAC can produce outstanding performance while making registration efficient.
\section{Conclusion}
\label{sec:conclusion}
In this paper, we presented MAC to solve PCR by using the maximal clique constraint to generate precise pose hypotheses from correspondences. Our method achieves state-of-the-art performance on all tested datasets and can adapt to deep-learned methods to boost their performance. \\
\noindent\textbf{Limitation.} As shown in Table~\ref{tab:evaluation_difficulty} and Table~\ref{tab:3dmatch5000}, MAC produces accurate hypotheses but may fail to find them. In the future, we plan to develop a more convincing hypothesis evaluation technique utilizing semantic information. \\
\noindent\textbf{Acknowledgments.} This work is supported in part by the National Natural Science Foundation of China (NFSC) (No.U19B2037 and 62002295), Shaanxi Provincial Key R\&D Program (No.2021KWZ-03), and the Fundamental Research Funds for the Central Universities (No.D5000220352).
{\small
\bibliographystyle{ieee_fullname}
\bibliography{egbib}

\begin{thebibliography}{10}\itemsep=-1pt

\bibitem{ao2021spinnet}
Sheng Ao, Qingyong Hu, Bo Yang, Andrew Markham, and Yulan Guo.
\newblock Spinnet: Learning a general surface descriptor for 3d point cloud
  registration.
\newblock In {\em Proceedings of the IEEE/CVF Conference on Computer Vision and
  Pattern Recognition}, pages 11753--11762, 2021.

\bibitem{aoki2019pointnetlk}
Yasuhiro Aoki, Hunter Goforth, Rangaprasad~Arun Srivatsan, and Simon Lucey.
\newblock Pointnetlk: Robust \& efficient point cloud registration using
  pointnet.
\newblock In {\em Proceedings of the IEEE/CVF Conference on Computer Vision and
  Pattern Recognition}, pages 7163--7172, 2019.

\bibitem{bai2021pointdsc}
Xuyang Bai, Zixin Luo, Lei Zhou, Hongkai Chen, Lei Li, Zeyu Hu, Hongbo Fu, and
  Chiew-Lan Tai.
\newblock Pointdsc: Robust point cloud registration using deep spatial
  consistency.
\newblock In {\em Proceedings of the IEEE Conference on Computer Vision and
  Pattern Recognition}, pages 15859--15869. IEEE, 2021.

\bibitem{bai2020d3feat}
Xuyang Bai, Zixin Luo, Lei Zhou, Hongbo Fu, Long Quan, and Chiew-Lan Tai.
\newblock D3feat: Joint learning of dense detection and description of 3d local
  features.
\newblock In {\em Proceedings of the IEEE/CVF Conference on Computer Vision and
  Pattern Recognition}, pages 6359--6367, 2020.

\bibitem{barath2018graph}
Daniel Barath and Ji{\v{r}}{\'\i} Matas.
\newblock Graph-cut ransac.
\newblock In {\em Proceedings of the IEEE Conference on Computer Vision and
  Pattern Recognition}, pages 6733--6741, 2018.

\bibitem{bustos2017guaranteed}
Alvaro~Parra Bustos and Tat-Jun Chin.
\newblock Guaranteed outlier removal for point cloud registration with
  correspondences.
\newblock {\em IEEE Transactions on Pattern Analysis and Machine Intelligence},
  40(12):2868--2882, 2017.

\bibitem{chen20073d}
Hui Chen and Bir Bhanu.
\newblock 3d free-form object recognition in range images using local surface
  patches.
\newblock {\em Pattern Recognition Letters}, 28(10):1252--1262, 2007.

\bibitem{chen2022sc2}
Zhi Chen, Kun Sun, Fan Yang, and Wenbing Tao.
\newblock Sc2-pcr: A second order spatial compatibility for efficient and
  robust point cloud registration.
\newblock In {\em Proceedings of the IEEE/CVF Conference on Computer Vision and
  Pattern Recognition}, pages 13221--13231, 2022.

\bibitem{choy2020deep}
Christopher Choy, Wei Dong, and Vladlen Koltun.
\newblock Deep global registration.
\newblock In {\em Proceedings of the IEEE Conference on Computer Vision and
  Pattern Recognition}, pages 2514--2523. IEEE, 2020.

\bibitem{choy2019fully}
Christopher Choy, Jaesik Park, and Vladlen Koltun.
\newblock Fully convolutional geometric features.
\newblock In {\em Proceedings of the IEEE/CVF International Conference on
  Computer Vision}, pages 8958--8966, 2019.

\bibitem{drost2010model}
Bertram Drost, Markus Ulrich, Nassir Navab, and Slobodan Ilic.
\newblock Model globally, match locally: Efficient and robust 3d object
  recognition.
\newblock In {\em IEEE Computer Society Conference on Computer Vision and
  Pattern Recognition}, pages 998--1005. IEEE, 2010.

\bibitem{eppstein2010listing}
David Eppstein, Maarten L{\"o}ffler, and Darren Strash.
\newblock Listing all maximal cliques in sparse graphs in near-optimal time.
\newblock In {\em International Symposium on Algorithms and Computation}, pages
  403--414. Springer, 2010.

\bibitem{fischler1981random}
Martin~A Fischler and Robert~C Bolles.
\newblock Random sample consensus: a paradigm for model fitting with
  applications to image analysis and automated cartography.
\newblock {\em Communications of the ACM}, 24(6):381--395, 1981.

\bibitem{fu2021robust}
Kexue Fu, Shaolei Liu, Xiaoyuan Luo, and Manning Wang.
\newblock Robust point cloud registration framework based on deep graph
  matching.
\newblock In {\em Proceedings of the IEEE/CVF Conference on Computer Vision and
  Pattern Recognition}, pages 8893--8902, 2021.

\bibitem{geiger2012we}
Andreas Geiger, Philip Lenz, and Raquel Urtasun.
\newblock Are we ready for autonomous driving? the kitti vision benchmark
  suite.
\newblock In {\em Proceedings of the IEEE Conference on Computer Vision and
  Pattern Recognition}, pages 3354--3361. IEEE, 2012.

\bibitem{gojcic2019perfect}
Zan Gojcic, Caifa Zhou, Jan~D Wegner, and Andreas Wieser.
\newblock The perfect match: 3d point cloud matching with smoothed densities.
\newblock In {\em Proceedings of the IEEE/CVF Conference on Computer Vision and
  Pattern Recognition}, pages 5545--5554, 2019.

\bibitem{guo20143d}
Yulan Guo, Mohammed Bennamoun, Ferdous Sohel, Min Lu, and Jianwei Wan.
\newblock 3d object recognition in cluttered scenes with local surface
  features: A survey.
\newblock {\em IEEE Transactions on Pattern Analysis and Machine Intelligence},
  36(11):2270--2287, 2014.

\bibitem{huang2021predator}
Shengyu Huang, Zan Gojcic, Mikhail Usvyatsov, Andreas Wieser, and Konrad
  Schindler.
\newblock Predator: Registration of 3d point clouds with low overlap.
\newblock In {\em Proceedings of the IEEE/CVF Conference on Computer Vision and
  Pattern Recognition}, pages 4267--4276, 2021.

\bibitem{lee2021deep}
Junha Lee, Seungwook Kim, Minsu Cho, and Jaesik Park.
\newblock Deep hough voting for robust global registration.
\newblock In {\em Proceedings of the IEEE/CVF International Conference on
  Computer Vision}, pages 15994--16003, 2021.

\bibitem{leordeanu2005spectral}
Marius Leordeanu and Martial Hebert.
\newblock A spectral technique for correspondence problems using pairwise
  constraints.
\newblock 2005.

\bibitem{li2021practical}
Jiayuan Li.
\newblock A practical o (n2) outlier removal method for point cloud
  registration.
\newblock {\em IEEE Transactions on Pattern Analysis and Machine Intelligence},
  2021.

\bibitem{li2022lepard}
Yang Li and Tatsuya Harada.
\newblock Lepard: Learning partial point cloud matching in rigid and deformable
  scenes.
\newblock In {\em Proceedings of the IEEE/CVF Conference on Computer Vision and
  Pattern Recognition}, pages 5554--5564, 2022.

\bibitem{lin2022planted}
Muyuan Lin, Varun Murali, and Sertac Karaman.
\newblock A planted clique perspective on hypothesis pruning.
\newblock {\em IEEE Robotics and Automation Letters}, 7(2):5167--5174, 2022.

\bibitem{lin2022k}
Yu-Kai Lin, Wen-Chieh Lin, and Chieh-Chih Wang.
\newblock K-closest points and maximum clique pruning for efficient and
  effective 3-d laser scan matching.
\newblock {\em IEEE Robotics and Automation Letters}, 7(2):1471--1477, 2022.

\bibitem{mian2005automatic}
Ajmal~S Mian, Mohammed Bennamoun, and Robyn~A Owens.
\newblock Automatic correspondence for 3d modeling: an extensive review.
\newblock {\em International Journal of Shape Modeling}, 11(02):253--291, 2005.

\bibitem{mian2006novel}
Ajmal~S Mian, Mohammed Bennamoun, and Robyn~A Owens.
\newblock A novel representation and feature matching algorithm for automatic
  pairwise registration of range images.
\newblock {\em International Journal of Computer Vision}, 66(1):19--40, 2006.

\bibitem{pais20203dregnet}
G~Dias Pais, Srikumar Ramalingam, Venu~Madhav Govindu, Jacinto~C Nascimento,
  Rama Chellappa, and Pedro Miraldo.
\newblock 3dregnet: A deep neural network for 3d point registration.
\newblock In {\em Proceedings of the IEEE Conference on Computer Vision and
  Pattern Recognition}, pages 7193--7203. IEEE, 2020.

\bibitem{parra2019practical}
Alvaro Parra, Tat-Jun Chin, Frank Neumann, Tobias Friedrich, and Maximilian
  Katzmann.
\newblock A practical maximum clique algorithm for matching with pairwise
  constraints.
\newblock {\em arXiv preprint arXiv:1902.01534}, 2019.

\bibitem{pomerleau2012challenging}
Fran{\c{c}}ois Pomerleau, Ming Liu, Francis Colas, and Roland Siegwart.
\newblock Challenging data sets for point cloud registration algorithms.
\newblock {\em The International Journal of Robotics Research},
  31(14):1705--1711, 2012.

\bibitem{qin2022geometric}
Zheng Qin, Hao Yu, Changjian Wang, Yulan Guo, Yuxing Peng, and Kai Xu.
\newblock Geometric transformer for fast and robust point cloud registration.
\newblock In {\em Proceedings of the IEEE/CVF Conference on Computer Vision and
  Pattern Recognition}, pages 11143--11152, 2022.

\bibitem{quan2020compatibility}
Siwen Quan and Jiaqi Yang.
\newblock Compatibility-guided sampling consensus for 3-d point cloud
  registration.
\newblock {\em IEEE Transactions on Geoscience and Remote Sensing},
  58(10):7380--7392, 2020.

\bibitem{rusu2009fast}
Radu~Bogdan Rusu, Nico Blodow, and Michael Beetz.
\newblock Fast point feature histograms (fpfh) for 3d registration.
\newblock In {\em IEEE International Conference on Robotics and Automation},
  pages 3212--3217. IEEE, 2009.

\bibitem{rusu20113d}
Radu~Bogdan Rusu and Steve Cousins.
\newblock 3d is here: Point cloud library (pcl).
\newblock In {\em IEEE International Conference on Robotics and Automation},
  pages 1--4. IEEE, 2011.

\bibitem{sipiran2011harris}
Ivan Sipiran and Benjamin Bustos.
\newblock Harris 3d: a robust extension of the harris operator for interest
  point detection on 3d meshes.
\newblock {\em The Visual Computer}, 27(11):963--976, 2011.

\bibitem{tombari2010unique}
Federico Tombari, Samuele Salti, and Luigi~Di Stefano.
\newblock Unique signatures of histograms for local surface description.
\newblock In {\em European Conference on Computer Vision}, pages 356--369.
  Springer, 2010.

\bibitem{wang2022you}
Haiping Wang, Yuan Liu, Zhen Dong, and Wenping Wang.
\newblock You only hypothesize once: Point cloud registration with
  rotation-equivariant descriptors.
\newblock In {\em Proceedings of the ACM International Conference on
  Multimedia}, pages 1630--1641, 2022.

\bibitem{yang2020teaser}
Heng Yang, Jingnan Shi, and Luca Carlone.
\newblock Teaser: Fast and certifiable point cloud registration.
\newblock {\em IEEE Transactions on Robotics}, 37(2):314--333, 2020.

\bibitem{yang2016fast}
Jiaqi Yang, Zhiguo Cao, and Qian Zhang.
\newblock A fast and robust local descriptor for 3d point cloud registration.
\newblock {\em Information Sciences}, 346:163--179, 2016.

\bibitem{yang2022correspondence}
Jiaqi Yang, Jiahao Chen, Siwen Quan, Wei Wang, and Yanning Zhang.
\newblock Correspondence selection with loose-tight geometric voting for 3d
  point cloud registration.
\newblock {\em IEEE Transactions on Geoscience and Remote Sensing}, 2022.

\bibitem{yang2021sac}
Jiaqi Yang, Zhiqiang Huang, Siwen Quan, Zhaoshuai Qi, and Yanning Zhang.
\newblock Sac-cot: Sample consensus by sampling compatibility triangles in
  graphs for 3-d point cloud registration.
\newblock {\em IEEE Transactions on Geoscience and Remote Sensing}, 60:1--15,
  2021.

\bibitem{yang2021toward}
Jiaqi Yang, Zhiqiang Huang, Siwen Quan, Qian Zhang, Yanning Zhang, and Zhiguo
  Cao.
\newblock Toward efficient and robust metrics for ransac hypotheses and 3d
  rigid registration.
\newblock {\em IEEE Transactions on Circuits and Systems for Video Technology},
  32(2):893--906, 2021.

\bibitem{yang2015go}
Jiaolong Yang, Hongdong Li, Dylan Campbell, and Yunde Jia.
\newblock Go-icp: A globally optimal solution to 3d icp point-set registration.
\newblock {\em IEEE Transactions on Pattern Analysis and Machine Intelligence},
  38(11):2241--2254, 2015.

\bibitem{yang2019ranking}
Jiaqi Yang, Yang Xiao, Zhiguo Cao, and Weidong Yang.
\newblock Ranking 3d feature correspondences via consistency voting.
\newblock {\em Pattern Recognition Letters}, 117:1--8, 2019.

\bibitem{yu2021cofinet}
Hao Yu, Fu Li, Mahdi Saleh, Benjamin Busam, and Slobodan Ilic.
\newblock Cofinet: Reliable coarse-to-fine correspondences for robust
  pointcloud registration.
\newblock {\em Advances in Neural Information Processing Systems}, 34, 2021.

\bibitem{zeng20173dmatch}
Andy Zeng, Shuran Song, Matthias Nie{\ss}ner, Matthew Fisher, Jianxiong Xiao,
  and Thomas Funkhouser.
\newblock 3dmatch: Learning local geometric descriptors from rgb-d
  reconstructions.
\newblock In {\em Proceedings of the IEEE Conference on Computer Vision and
  Pattern Recognition}, pages 1802--1811, 2017.

\bibitem{zhou2016fast}
Qian-Yi Zhou, Jaesik Park, and Vladlen Koltun.
\newblock Fast global registration.
\newblock In {\em European Conference on Computer Vision}, pages 766--782.
  Springer, 2016.

\end{thebibliography}
}
\clearpage
\begin{center}
    \Large
    3D Registration with Maximal Cliques\\ \emph{Supplementary Material}
    \\[10pt]
\end{center}
\setcounter{section}{0}
\renewcommand\thesection{\Alph{section}}

\section{Comparison of FOG and SOG}
\begin{figure}[h]
    \centering
    \includegraphics[width=\linewidth]{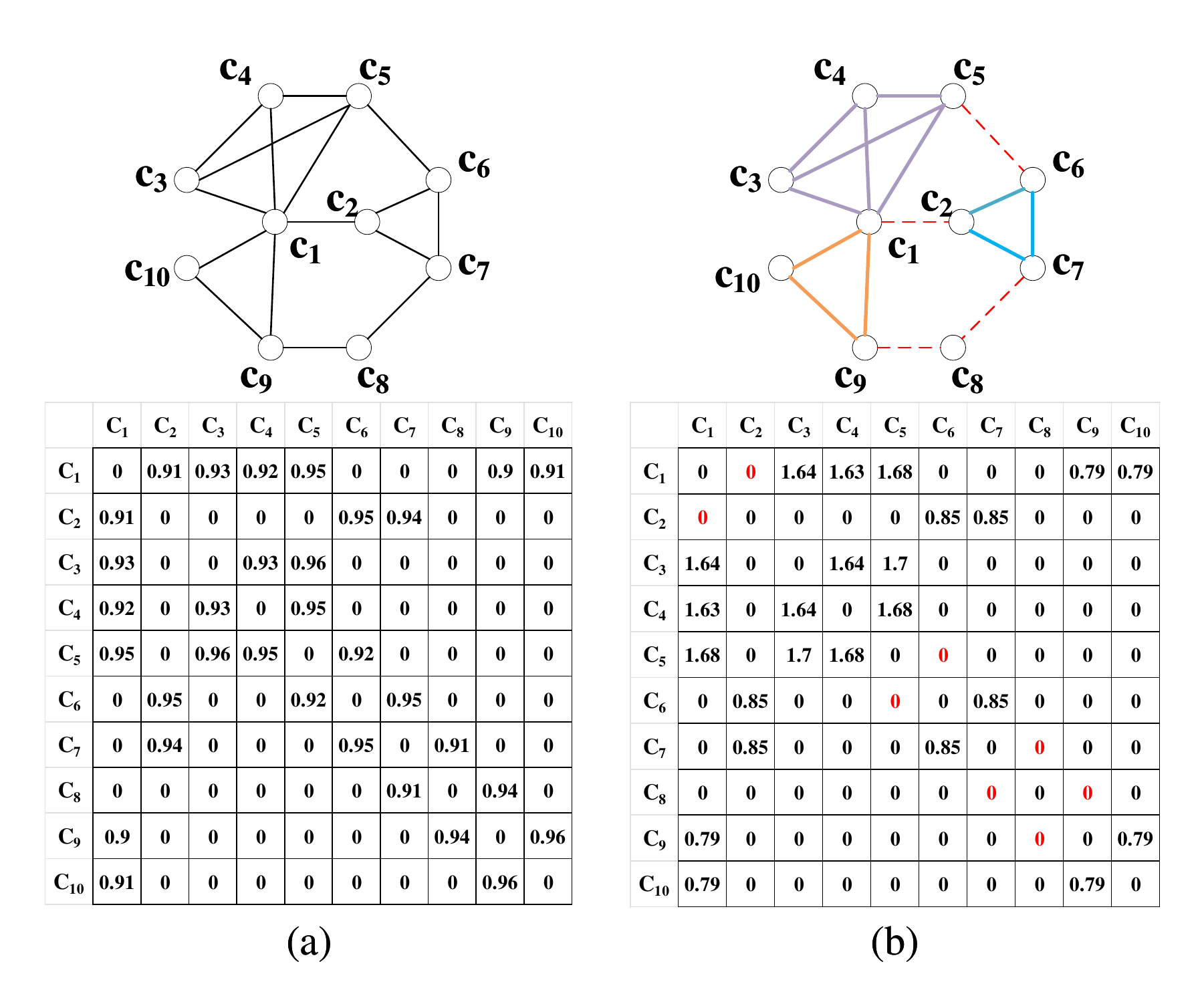}
    \caption{An example that illustrates the relationship between FOG and SOG. $\bf{(a)}$ FOG and its weight matrix. $\bf{(b)}$ SOG and its weight matrix.}
\label{fig:SC2}
\end{figure}

\begin{table*}[b]
      \centering
      \resizebox{\linewidth}{!}{
        \begin{tabular}{ccccc}
\hline
Dataset   & Data type     & Nuisances                       & Application scenario & \# Matching pairs \\ \hline
U3M~\cite{mian2006novel}               & Object        & Limited overlap, self-occlusion & Registration         & 496              \\
3DMatch~\cite{zeng20173dmatch}       & Indoor scene  &          Occlusion, real noise       &     Registration      & 1623             \\
3DLoMatch~\cite{huang2021predator}               & Indoor scene  &   Limited overlap, occlusion, real noise            &   Registration     & 1781             \\
KITTI~\cite{geiger2012we}              & Outdoor scene &   Clutter, occlusion, real noise   &   Detection, registration,  segmentation                   & 555              \\
ETH~\cite{pomerleau2012challenging}                     & Outdoor scene &  Limited overlap, clutter, occlusion, real noise   &    Feature description, registration       & 713              \\ \hline
\end{tabular}}
        \caption{Information of all tested datasets.}
      \label{tab:scene}
    \end{table*}
As shown in Fig.~\ref{fig:SC2}: 1) SOG considers the commonly compatible matches in the global set of the matched pairs rather than only the geometric consistency, making it more consistent and more robust in the case of high outlier rates; 2) SOG is sparser than FOG, and therefore beneficial in making the search of cliques more rapid.

The weights of the edge $e_{ij}=({\textbf{c}_i},{\textbf{c}_j})$ in the FOG are transformed as follows to obtain the corresponding second-order weights:
\begin{equation}
    {w_{SOG}}({e_{ij}}) = {w_{FOG}}({e_{ij}}) \cdot \sum\limits_{\scriptstyle{e_{ik}} \in {\mathbf E}\hfill\atop
\scriptstyle{e_{jk}} \in {\mathbf E}\hfill} {\left[ {{w_{FOG}}({e_{ik}}) \cdot {w_{FOG}}({e_{jk}})} \right]}.
\end{equation}
If no remaining nodes form edges with both ${\textbf{c}_i}$ and ${\textbf{c}_j}$, ${w_{SOG}}({e_{ij}})$ will be 0, which demonstrates that ${e_{ij}}$ will be removed from SOG then. In Fig.~\ref{fig:SC2}(b), the four edges ${e_{12}}$, ${e_{56}}$, ${e_{78}}$ and ${e_{89}}$ are removed, and the whole graph is divided into subgraphs that contain several cliques naturally.

\section{Additional Experiments}
The information of all tested datasets is presented in Table~\ref{tab:scene}.

\noindent\textbf{Results on ETH.} Additionally, we also test our method on the outdoor dataset ETH~\cite{pomerleau2012challenging}, which contains more complex geometries compared with 3DMatch~\cite{zeng20173dmatch}. FPFH~\cite{rusu2009fast}, FCGF~\cite{choy2019fully}, and Spinnet~\cite{ao2021spinnet} are employed to generate correspondences, from which registration will then be performed by RANSAC-50K and MAC. The number of sampled points or correspondences is set to 5000. Registration is considered successful when the RE $\leq$ 15\textdegree and TE $\leq$ 30 cm. The quality of generated correspondence and registration results are reported in Table~\ref{tab:ETH_inlier} and Table~\ref{tab:ETH}, respectively.

\begin{table}[h]
\centering
      \resizebox{\linewidth}{!}{
\begin{tabular}{c|cc|cc|c}
\hline
\multirow{2}{*}{} & \multicolumn{2}{c|}{\textbf{Gazebo}} & \multicolumn{2}{c|}{\textbf{Wood}} & \multirow{2}{*}{Avg.} \\
                  & Summer            & Winter           & Autumn           & Summer          &                       \\ \hline
FPFH~\cite{rusu2009fast}              &     0.42     &   0.24    &     0.21   &   0.26  &  0.29            \\
FCGF~\cite{choy2019fully}             &    2.34    &  1.25  &     1.35  &      1.68   &   1.62          \\
Spinnet~\cite{ao2021spinnet}           &    16.67    &  13.73     &    12.20       &  14.67   &  14.40             \\ \hline
\end{tabular}}
\caption{Inlier ratio (\%) of generated correspondence on ETH dataset. }
\label{tab:ETH_inlier}
\end{table}
\begin{table}[h]
\centering
      \resizebox{\linewidth}{!}{
\begin{tabular}{c|cc|cc|c}
\hline
\multirow{2}{*}{} & \multicolumn{2}{c|}{\textbf{Gazebo}} & \multicolumn{2}{c|}{\textbf{Wood}} & \multirow{2}{*}{Avg.} \\
                  & Summer            & Winter           & Autumn           & Summer          &                       \\ \hline
FPFH\cite{rusu2009fast}              &          16.85         &  10.03               &      10.43          &   10.40           &    11.92             \\
FCGF\cite{choy2019fully}             &      54.35            &        28.03      &     52.17     &     51.20     &    42.78           \\
Spinnet\cite{ao2021spinnet}           &    98.37       &     83.05       &  100.00           &     99.20    &  92.57              \\ \hline
\multirow{2}{*}{FPFH+MAC}          &    46.74      &   27.68     &      33.04            &      43.20           &      36.12               \\
& 29.89$\uparrow$ & 17.65$\uparrow$ & 
 22.61$\uparrow$ & 32.80$\uparrow$ & 24.20$\uparrow$\\ \hdashline
\multirow{2}{*}{FCGF+MAC}          &   75.54   &      42.91    &     71.30             &       73.60          &         61.29              \\
    & 21.19$\uparrow$ & 14.88$\uparrow$ & 19.13$\uparrow$ & 22.40$\uparrow$ & 18.51$\uparrow$\\ \hdashline
\multirow{2}{*}{Spinnet+MAC}       &      98.91      &    87.54              &        100.00    &       100.00          &        94.67      \\ 
& 0.54$\uparrow$ & 4.49$\uparrow$ & - & 0.80$\uparrow$ & 2.10$\uparrow$ \\\hline
\end{tabular}}
\caption{Registration recall (\%) boosting for various descriptors combined with MAC on ETH dataset. }
\label{tab:ETH}
\end{table}

The results suggest that when a defect in a descriptor leads to a very low inlier rate for generating the correspondence set, MAC is still effective in finding the accurate consistent subset from it, thus greatly boosting the registration recall. The registration recall obtained by using MAC is 24.2\% higher than RANSAC when combined with FPFH, and 18.51\% higher when combined with FCGF. MAC working with Spinnet achieves a registration recall of {\bf{94.67\%}} on ETH.

\noindent\textbf{Time and memory analysis.} Efficiency  and memory consumption results of several well-performed methods are shown in Tables \ref{tab:time_table} and \ref{tab:memory_table}, respectively. Regarding efficiency experiments, all methods have been tested for ten rounds, and the mean and standard deviation results are reported.
All methods were executed in the CPU. The results indicate that MAC is quite lightweight and efficient when the input correspondence number is less than 2.5k.

\begin{table}[h]
\centering
\resizebox{\linewidth}{!}{
\begin{tabular}{l|ccccc}
\hline
\# Corr.        & 250   & 500   & 1000   & 2500    & 5000    \\ \hline
PointDSC                  & 32.24$\pm$0.81 & 78.38$\pm$0.89 & 240.46$\pm$2.18 & 1401.97$\pm$12.24 & 5504.11$\pm$10.32 \\
TEASER++                  & 6.40$\pm$1.88  & 6.68$\pm$0.66  & 16.74$\pm$1.21  & 104.24$\pm$0.53  & 484.93$\pm$1.87  \\
$\rm{SC}^2$-PCR & 19.34$\pm$0.63 & 63.23$\pm$0.55 & 215.98$\pm$1.24 & 1282.73$\pm$4.05 & 5210.17$\pm$8.30 \\ \hline
MAC                       & 7.32$\pm$0.55  & 23.32$\pm$0.38 & 56.45$\pm$1.41  & 282.67$\pm$7.83  & 3259.38$\pm$12.66 \\ \hline
\end{tabular}}
\caption{Comparisons on average time consumption (ms).}
\label{tab:time_table}
\end{table}

\begin{table}[h]
\centering
\resizebox{\linewidth}{!}{
\begin{tabular}{l|ccccc}
\hline
\# Corr.        & 250 & 500 & 1000 & 2500 & 5000 \\ \hline
PointDSC                  &   3531.46  & 3538.26    &  3582.57    &   3634.22   &  3736.10    \\
TEASER++                  &   1631.92  &  1634.77   &     2029.22  &   2266.84  &   2484.83   \\
$\rm{SC}^2$-PCR &  448.01   &  453.18   &    508.40  &   621.27   &   690.22   \\ \hline
MAC                       &   15.59  &  17.43   &  23.49    &   52.79   &   150.86   \\ \hline
\end{tabular}}
\caption{Comparisons on average memory consumption (MB).}
\label{tab:memory_table}
\end{table}
\section{Visualizations}
\begin{figure*}[hb]
    \centering
    \includegraphics[width=\linewidth]{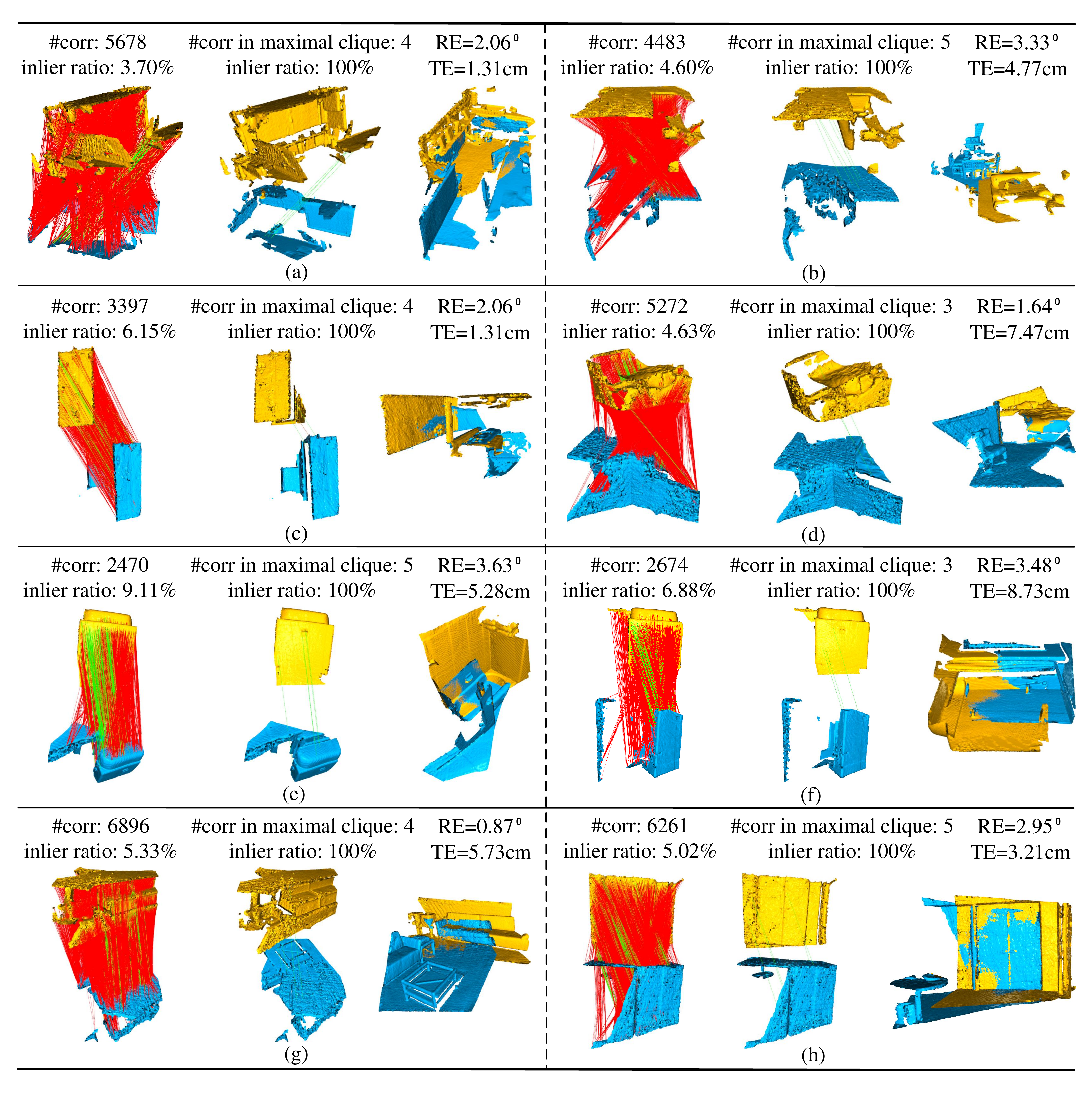}
    \caption{Registration process-visualizations of MAC on 3DMatch.}
\label{fig:vis_mac}
\end{figure*}

\begin{figure*}
    \centering
    \includegraphics[width=0.92\linewidth]{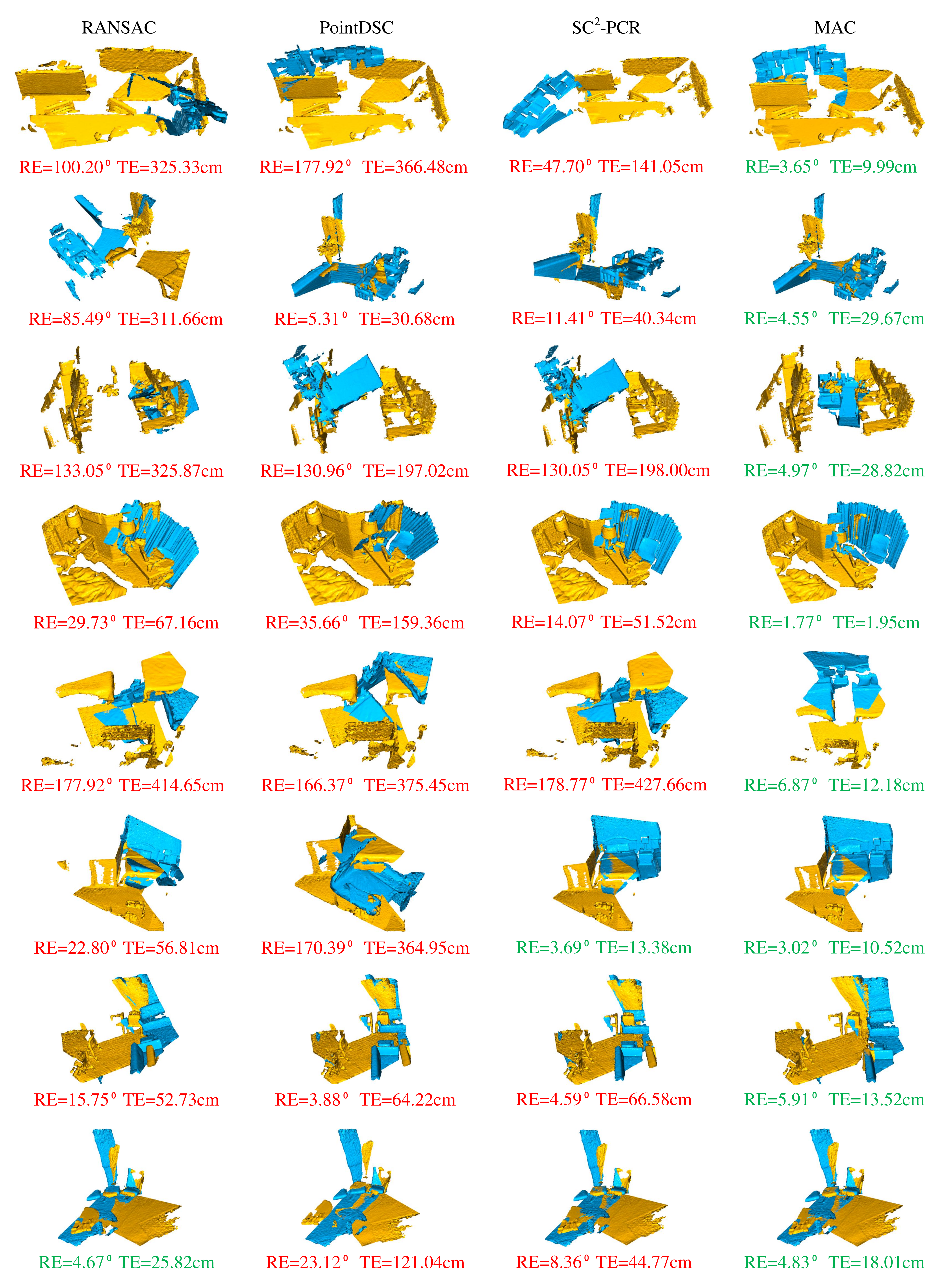}
    \caption{Qualitative comparison on 3DLoMatch. \textcolor{red}{Red} and \textcolor{green}{green} represent failed and successful registration, respectively.}
    \label{fig:cmp_on_3dmatch}
\end{figure*}

\begin{figure*}
    \centering
    \includegraphics[width=0.9\linewidth]{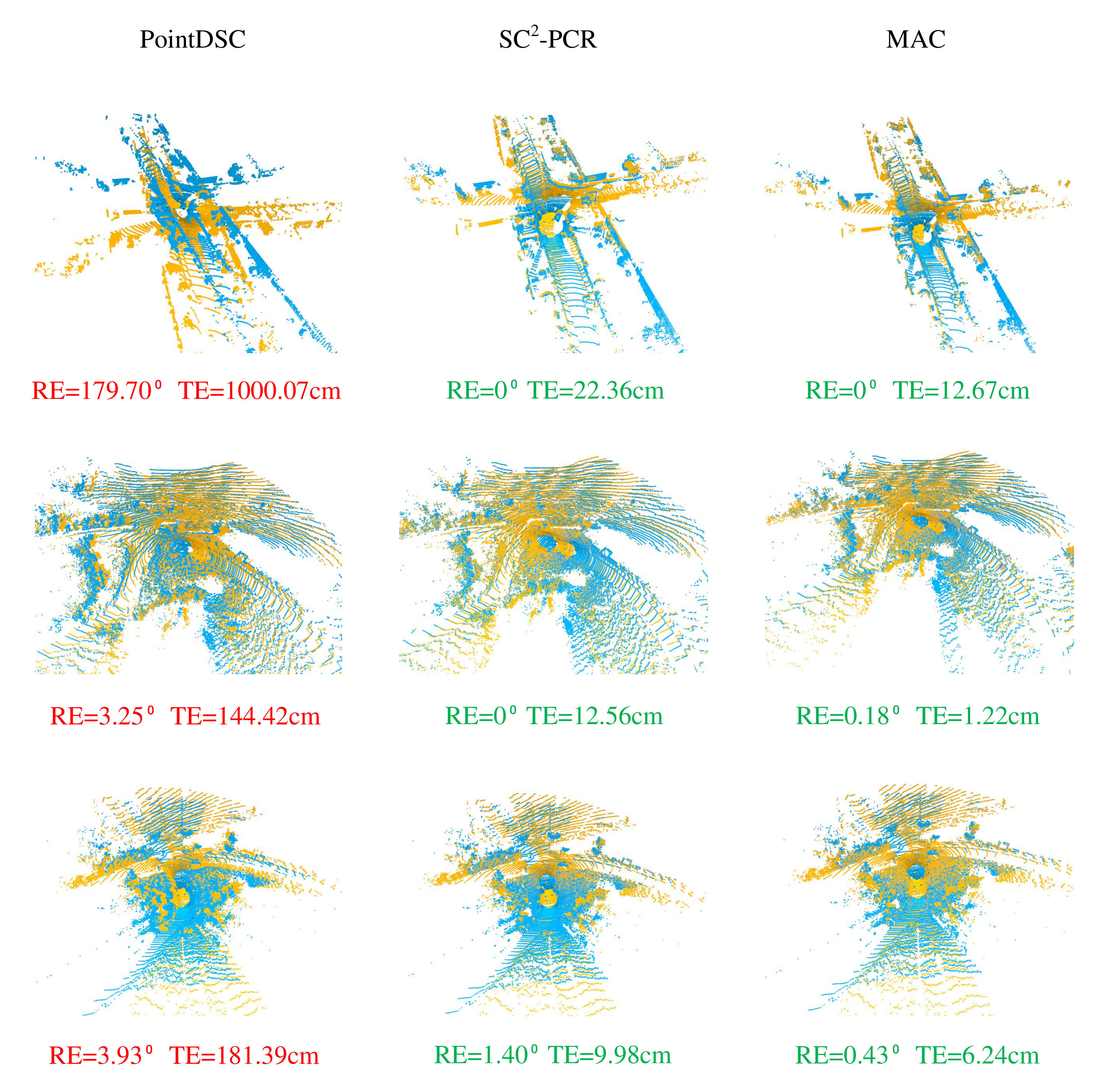}
    \caption{Qualitative comparison on KITTI.}
    \label{fig:kitti}
\end{figure*}

\begin{figure*}
    \centering
    \includegraphics[width=0.85\linewidth]{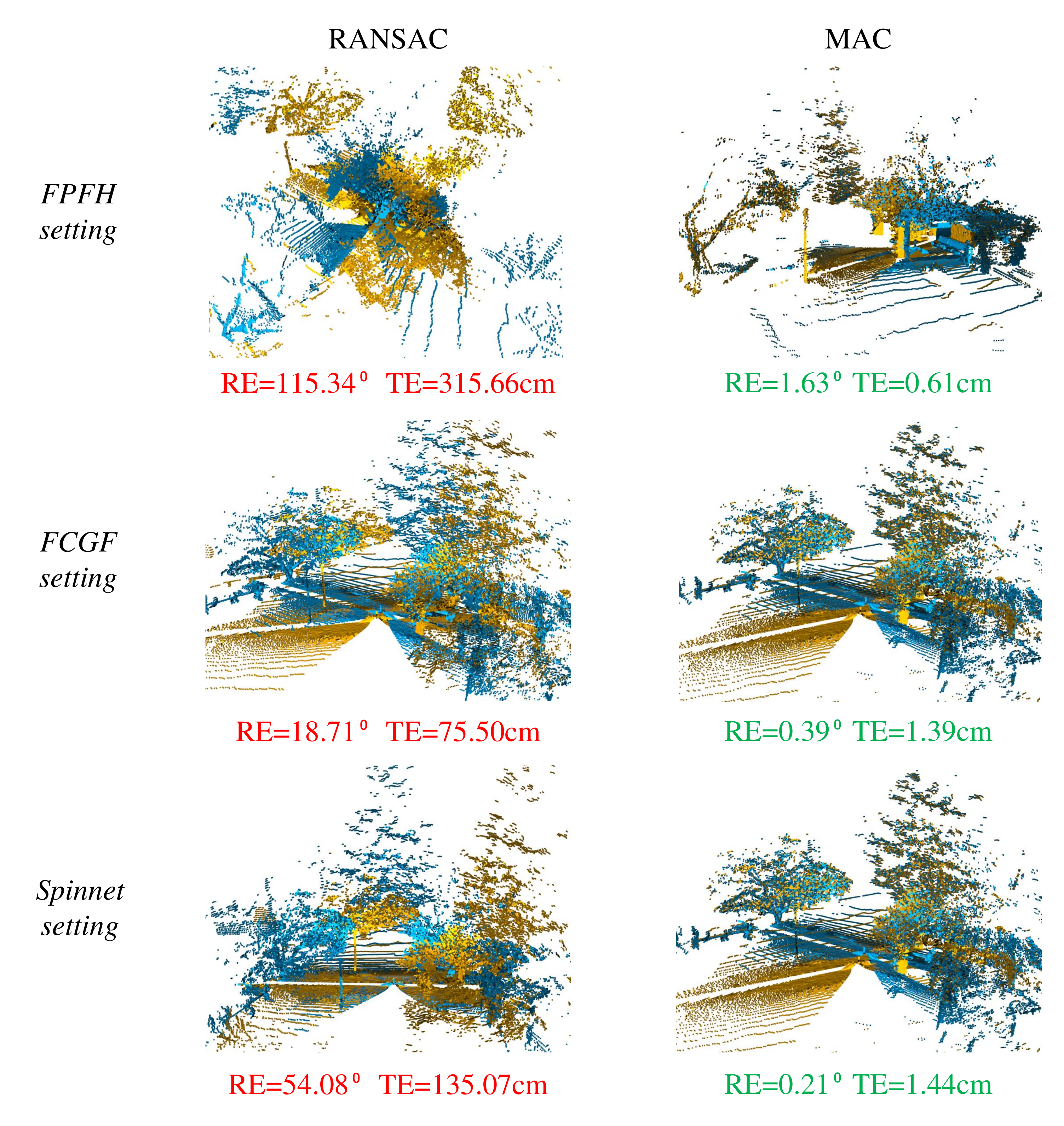}
    \caption{Qualitative comparison on ETH.}
    \label{fig:ETH}
\end{figure*}

We show more registration results in Figs.~\ref{fig:vis_mac}-\ref{fig:ETH}.

\clearpage

\clearpage

\clearpage


\end{document}